\pdfoutput=1
\documentclass[10pt,twocolumn,letterpaper]{article}
\usepackage{lwpaper}

\usepackage{iftex}
\ifXeTeX
  \usepackage{xeCJK}
  \newcommand{\zh}[1]{#1}
\else
  \usepackage{CJKutf8}
  \newcommand{\zh}[1]{\begin{CJK}{UTF8}{gbsn}#1\end{CJK}}
\fi

\usepackage{cuted} %

\usepackage{colortbl}
\definecolor{homebg}{HTML}{FCE4CC}
\definecolor{oursbg}{HTML}{EEF3F9}
\definecolor{navgood}{HTML}{D6EEDA}  %
\definecolor{navbad}{HTML}{F8D5D1}   %

\usepackage{arydshln}
\newcommand{\dashmidrule}{\noalign{\vskip\aboverulesep}\hdashline\noalign{\vskip\belowrulesep}}

\usepackage{currfile} %

\usepackage{caption} %

\usepackage{dblfloatfix}
\usepackage{placeins}
\definecolor{linkblue}{rgb}{0.21,0.49,0.74}
\usepackage[pagebackref,breaklinks,colorlinks,allcolors=linkblue]{hyperref}
\hypersetup{pdftitle={LaWanle: Why Can't Leading Driving Models Make a Simple Turn?},
  pdfauthor={Chengzhi Gao, Junzhe Gao, Yongzhi Liu, Weichao Zhuang, Weilong Hu, Taonian Liang, Guodong Yin}}

\title{LaWanle:\thanks{\emph{LaWanle} (\zh{拉完了}): Chinese slang for ``utterly awful''.}\hspace{0.55em}Why Can't Leading Driving Models\\Make a Simple Turn?}

\author{%
Chengzhi Gao$^{1}$\quad
Junzhe Gao$^{1}$\quad
Yongzhi Liu$^{1}$\quad
Weichao Zhuang$^{1,\dagger}$\\
Weilong Hu$^{2}$\quad
Taonian Liang$^{2}$\quad
Guodong Yin$^{1}$\\[0.5em]
{\normalsize $^{1}$Southeast University\qquad $^{2}$Chery International}\\[0.2em]
{\small $^{\dagger}$Corresponding author: \texttt{wezhuang@seu.edu.cn}}}

\begin{document}
\addtocontents{toc}{\protect\setcounter{tocdepth}{-1}}
\maketitle
\begin{abstract}
We were surprised to find that leading end-to-end driving models across major paradigms, despite scoring close to human drivers on NAVSIM, cannot reliably make a simple 90-degree turn in photorealistic closed-loop simulators.
To investigate why, we build LaWanle, a benchmark of turns drawn from three such simulators.
Driving each turn from several starts, we find that the same model passes it from one start and fails it from another, and that the two drives look alike until they enter the junction.
Rotating the car by a few degrees at the junction entry can fail a turn it otherwise passes.
Conversely, a slight heading correction to the plans can rescue about half of the failed turns we tested.
Such errors arise when a model drives by itself, so we post-train an open-source lane-keeping model with closed-loop DAgger on its lateral path.
Training 5\,\% of its parameters is enough to raise its turn pass rate from 19\,\% to 76\,\%.
The gain transfers to the two other benchmarks, where the policy also scores among the leading models.
Further fine-tuning on recovery data repairs the deviations it is trained on but, against a control trained equally long, improves turns on none of the three benchmarks, and every variant stays far below a human driver.
\end{abstract}
\section{Introduction}
\label{sec:intro}

\renewcommand{\topfraction}{0.95}\renewcommand{\textfraction}{0.05}
\begin{figure}[!t]
  \centering
  \includegraphics[width=\linewidth]{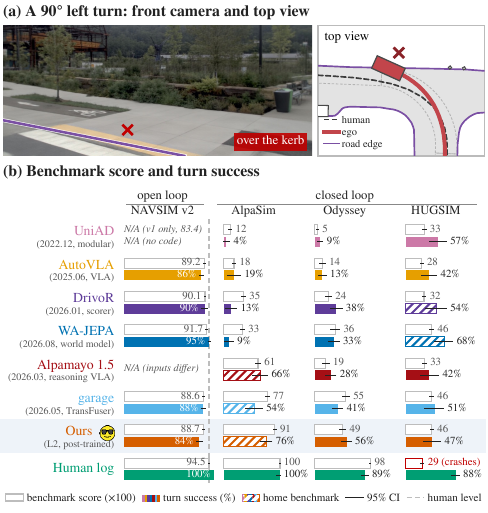}
  \caption{\textbf{Models that turn on NAVSIM fail turns in closed loop.}
  (a)~An example: a failed left turn on Odyssey.
  (b)~Benchmark score and turn success of seven models and the recorded human drive (Supp.~Sec.~\ref{sup:bench-lawanle}); turn success is the share of plans that pass a strict rule on NAVSIM and
  the share of turns passed under a relaxed rule in closed loop.}
  \label{fig:teaser}
\end{figure}
\begin{figure*}[t]
\centering
\includegraphics[width=\textwidth]{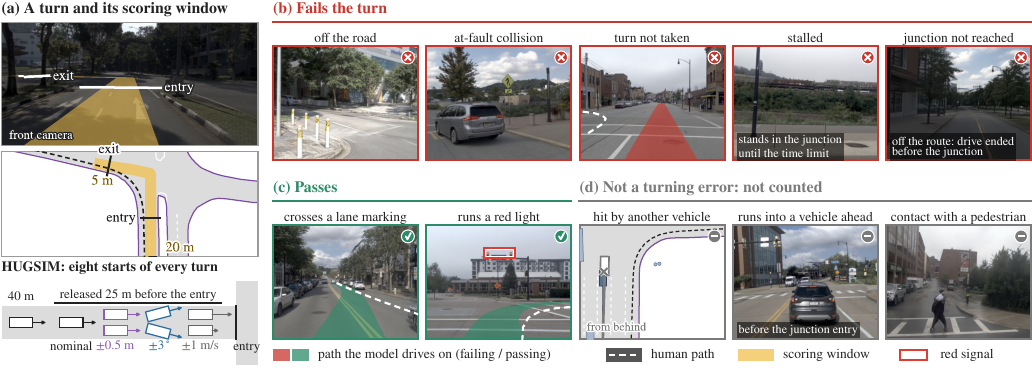}
\caption{\textbf{What LaWanle scores.}
(a)~A turn in the front camera and from above: it is scored from 20\,m before the junction entry to 5\,m past the exit, and on HUGSIM it is driven from eight starts.
(b)~The five failures of a turn.
(c)~Events that pass.
(d)~Events that are not turning errors: the turn is removed from the denominator, or the contact is ignored and the turn is scored on what follows.}
\label{fig:exam}
\end{figure*}

End-to-end driving models are now developed and compared mainly on NAVSIM~\cite{navsim,pseudosim}, which scores a 4\,s plan made from one logged frame and, in its second version, a second plan from views pre-rendered around the logged position 4\,s later.
Since HUGSIM~\cite{hugsim}, photorealistic closed-loop benchmarks such as AlpaSim~\cite{alpasim} and Odyssey~\cite{odyssey} have let a model drive for 20--100\,s in scenes reconstructed from real driving data, as it would on a real road.

We tested a wide range of leading end-to-end driving models across paradigms on these benchmarks, and we were surprised to find that they still cannot reliably complete a \mbox{basic} turn.
We select six of them as representatives of their paradigms~\cite{uniad,autovla,drivor,garage,wa_jepa,alpamayo_r1}.
On NAVSIM, under a strict rule, their turn success rate is as high as 86--95\,\% (the four with a NAVSIM score; Fig.~\ref{fig:teaser}; Supp.~Sec.~\ref{sup:bench-navsim}).
In closed loop, under a relaxed rule, they pass on average 36\,\% of the turns, and the best of them 68\,\%.
When a turn fails, the car most often leaves the road or runs into the static scene beside it; at-fault collisions with other vehicles account for about 1/10 of the failures.

Why has this gone unnoticed?
NAVSIM keeps its test logs out of training, but every state it tests is one the human driver reached: its frames lie on the recorded drive, and the second stage of its second version renders views near the position of the driver 4\,s later.
These are the states imitation learning trains on, and no error of the model moves it out of them.
Its top is crowded: leading models score within a few points of each other and approach or surpass the human driver~\cite{drivor,simscale,drivezero}.
Of the closed-loop benchmarks, AlpaSim and Odyssey are recent, and few models report on them so far~\cite{opted,eco,openlongtail}.
Most closed-loop results are on HUGSIM, where the best published zero-shot score is below half of the full score.
Papers that report on HUGSIM give this one score, split at most by difficulty tier or source dataset, and show turns only in example figures.
That score mostly measures the reaction to scripted traffic: 81\,\% of the scenarios insert vehicles, and in a quarter of all scenarios one of them is steered to collide with the car.
The recorded human drive, which does not react to them, scores 0.88 without them and 0.15 with them (Supp.~Sec.~\ref{sup:bench-lawanle}).
A failed turn costs little in it: 10--16\,\% of the failed episodes end at a turn, and they carry 8--12\,\% of the lost score.
garage and WA-JEPA both score 0.46 but pass 51\,\% versus 68\,\% of the turns (Supp.~Sec.~\ref{sup:boards}).

We therefore take the turn out of the benchmark score and test it alone: LaWanle is a benchmark of 269 turns drawn from HUGSIM, AlpaSim and Odyssey (Fig.~\ref{fig:exam}, Sec.~\ref{sec:setup}).
On HUGSIM every turn is driven from eight slightly different starts.
The results show that the same model passes a turn from one start and fails it from another, and that the approach does not tell the two apart: 10\,m before the junction entry most starts head within 2\textdegree{} of the passing ones, and many of them still fail (Fig.~\ref{fig:arrival}).
They part inside the junction, where the heading of the failing start drifts away from that of the passing one.
A small heading deviation there can fail a turn: rotating the car by 4\textdegree{} at the junction entry fails a fifth to a third of the turns that a model passes without it, while shifting it by 5\,cm has almost no effect (Table~\ref{tab:interv}a).
A small correction can rescue one: rotating each plan by the heading error of the car, less than 2\textdegree{} in the median, rescues half of the failed turns, and the same rotation with a random sign rescues almost none (Table~\ref{tab:interv}b).

To study turning in isolation, we take the open-source lane-keeping model of openpilot~\cite{openpilot}, which is not designed for such turns, freeze its vision encoder and post-train the layers behind it, 5\,\% of its parameters, first on driving logs and then with closed-loop DAgger~\cite{dagger} in AlpaSim.
Its turn pass rate on AlpaSim rises from 19\,\% to 51\,\% after the offline stage and to 76\,\% after the closed-loop stage (Fig.~\ref{fig:training}).
On Odyssey and HUGSIM, where no stage is trained, it rises from 11\,\% to 56\,\% and from 3\,\% to 47\,\%.
The samples labelled from the states of the policy carry part of it: without them the pass rate falls by 9, 14 and 15 points on the three benchmarks.

The policy still fails a quarter to a half of the turns, and the analysis above points to small heading deviations inside the junction.
We therefore fine-tune it on data aimed at these deviations.
Recovery data~\cite{chauffeurnet,codevilla2019} push the car off its line at a turn and label the way back. They rescue 23 of 27 injected deviations on AlpaSim, against 14 for a control trained equally long with no new data, yet improve the turns the policy drives by itself on none of the three benchmarks.
With the navigation route perturbed in the same data, two fine-tunes run from all three seeds each pass more Odyssey turns than the seeds they start from, by 0.08 and 0.09 of the turns, and neither passes more HUGSIM turns.
A control that continues the training equally long with no new data explains most of this gain: it passes 0.09 more of the Odyssey turns than a control trained half as long, and against it no fine-tune, with or without a perturbed route, passes resolvably more turns on any benchmark.
For the first of the two fine-tunes the scores of the three benchmarks do not change resolvably, and its NAVSIM v2 EPDMS falls by 0.35.

In summary, our contributions are:
\begin{itemize}
  \item \textbf{Benchmark}: we build LaWanle, a benchmark that tests whether a model can turn. It holds 269 turns from three closed-loop simulators reconstructed from real driving logs and scores each turn on its own. We report on it six released models, our policy and the recorded human drive.
  \item \textbf{Analysis}: we examine why end-to-end driving models fail turns in closed loop. We compare paired starts of the same turn, intervene on the car at the turn entry and test the common explanations of closed-loop failure.
  \item \textbf{Training}: we post-train 5\,\% of the parameters of a lane-keeping model so that it turns on the benchmark of its training and on two others, and show with ablations and fine-tuning experiments which data make the gain transfer.
\end{itemize}
\section{Related Work}
\label{sec:related}

\begin{table*}[t]
\centering
\footnotesize\setlength{\tabcolsep}{2.6pt}
\begin{tabular}{@{}l@{\hspace{7pt}}c@{\hspace{7pt}}r@{\hspace{3pt}}lrrrrr@{\hspace{7pt}}r@{\hspace{3pt}}lrrrrr@{\hspace{7pt}}r@{\hspace{3pt}}lrrrrr@{}}
\toprule
 & {\color{gray}NAVSIM} & \multicolumn{7}{c}{AlpaSim (57 turns)} & \multicolumn{7}{c}{Odyssey (113 turns)} & \multicolumn{7}{c}{HUGSIM (99 $\times$ 8 starts)} \\
\cmidrule(l{2pt}r{2pt}){3-9}\cmidrule(l{2pt}r{2pt}){10-16}\cmidrule(l{2pt}r{2pt}){17-23}
Model & {\scriptsize\color{gray}PDMS/EPDMS} & \multicolumn{2}{c}{Pass (\%) $\uparrow$} & OR & CO & MT & ST & NR & \multicolumn{2}{c}{Pass (\%) $\uparrow$} & OR & CO & MT & ST & NR & \multicolumn{2}{c}{Pass (\%) $\uparrow$} & OR & CO & MT & ST & NR \\
\midrule
UniAD~\cite{uniad} & {\color{gray}83.4$^\dagger$/--} & 4.3 & {\scriptsize\color{gray}[0,\,11]} & 50 & 9 & 2 & 0 & 35 & 9.2 & {\scriptsize\color{gray}[4,\,15]} & 52 & 4 & 3 & 0 & 32 & 57.1 & {\scriptsize\color{gray}[42,\,72]} & 36 & 5 & 0 & 0 & 1 \\
AutoVLA~\cite{autovla} & {\color{gray}89.5/89.2} & 18.6 & {\scriptsize\color{gray}[7,\,30]} & 40 & 7 & 9 & 9 & 16 & 12.9 & {\scriptsize\color{gray}[6,\,20]} & 38 & 5 & 2 & 3 & 40 & 41.9 & {\scriptsize\color{gray}[28,\,55]} & 43 & 6 & 7 & 0 & 2 \\
DrivoR~\cite{drivor} & {\color{gray}93.7/90.1} & 13.2 & {\scriptsize\color{gray}[3,\,24]} & 18 & 18 & 21 & 13 & 16 & 37.8 & {\scriptsize\color{gray}[27,\,49]} & 19 & 11 & 14 & 1 & 16 & \cellcolor{homebg}54.3 & \cellcolor{homebg}{\scriptsize\color{gray}[35,\,72]} & 14 & 6 & 22 & 1 & 3 \\
WA-JEPA~\cite{wa_jepa} & {\color{gray}91.7/91.7} & 8.9 & {\scriptsize\color{gray}[2,\,18]} & 31 & 2 & 42 & 4 & 11 & 33.3 & {\scriptsize\color{gray}[24,\,43]} & 19 & 10 & 11 & 5 & 22 & \cellcolor{homebg}\textbf{68.1} & \cellcolor{homebg}{\scriptsize\color{gray}[51,\,83]} & 6 & 3 & 20 & 0 & 2 \\
Alpamayo 1.5~\cite{alpamayo_r1} & {\color{gray}--} & \cellcolor{homebg}66.0 & \cellcolor{homebg}{\scriptsize\color{gray}[52,\,78]} & 8 & 2 & 6 & 16 & 2 & 28.2 & {\scriptsize\color{gray}[19,\,38]} & 17 & 10 & 16 & 1 & 28 & 41.9 & {\scriptsize\color{gray}[25,\,58]} & 17 & 2 & 6 & 30 & 3 \\
garage~\cite{garage} & {\color{gray}88.6/88.6} & \cellcolor{homebg}54.2 & \cellcolor{homebg}{\scriptsize\color{gray}[40,\,69]} & 17 & 4 & 0 & 17 & 8 & 40.6 & {\scriptsize\color{gray}[32,\,50]} & 26 & 5 & 0 & 12 & 16 & 51.3$^\S$ & {\scriptsize\color{gray}[34,\,67]} & 22 & 5 & 3 & 17 & 3 \\
\rowcolor{oursbg}Ours & {\color{gray}88.5/88.7} & \cellcolor{homebg}\textbf{75.9} & \cellcolor{homebg}{\scriptsize\color{gray}[65,\,86]} & 7 & 10 & 0 & 6 & 1 & \textbf{56.4} & {\scriptsize\color{gray}[47,\,65]} & 20 & 22 & 0 & 1 & 1 & 46.6 & {\scriptsize\color{gray}[31,\,62]} & 45 & 5 & 2 & 0 & 1 \\
\midrule
\textit{Human driver} & {\color{gray}94.6/94.5} & 100 & {\scriptsize\color{gray}[94,\,100]} & 0 & 0 & 0 & 0 & 0 & 89.4 & {\scriptsize\color{gray}[83,\,95]} & 11 & 0 & 0 & 0 & 0 & 88.0 & {\scriptsize\color{gray}[80,\,95]} & 0 & 12 & 0 & 0 & 0 \\
\bottomrule
\end{tabular}
\caption{\textbf{LaWanle results.}
Pass rate (\%) of the scored turns with its 95\,\% interval, and the share of the scored turns (\%) that ends in each kind of failure: OR off the road, CO at-fault collision, MT turn not taken, ST stalled, NR junction not reached.
Bold: highest pass rate on a benchmark. {\setlength{\fboxsep}{0.8pt}\colorbox{homebg}{Shaded}}: home benchmark of the model, on which its authors report or train.
NAVSIM: v1 PDMS / v2 EPDMS of the checkpoint we drive ($^\dagger$a NAVSIM-trained checkpoint, from~\cite{navsim}; --: no score). $^\S$Fed the recorded route. Human driver: the recorded drive, on HUGSIM from the nominal start only.
Numbers and notes in full: Supp.~Sec.~\ref{sup:res-table}.}
\label{tab:exam}
\end{table*}

\paragraph{Open-loop and closed-loop evaluation.}
The gap between open-loop and closed-loop evaluation of driving models is well documented.
Displacement errors on nuScenes reward a model for extrapolating its ego status~\cite{admlp,egostatus}, and open-loop and closed-loop scores rank planners differently on nuPlan~\cite{parting}.
NAVSIM~\cite{navsim,pseudosim} replaced displacement errors with simulation-based metrics, and still scores each plan on its own: its second version adds plans from views rendered in advance near the path of the human driver, so a model never sees a view that its own plan produced.
We leave out the closed-loop benchmarks in CARLA~\cite{carla,bench2drive}, whose game-engine images and few scripted scenes let a policy overfit to the simulator~\cite{plant2}.
Closed-loop benchmarks in scenes reconstructed from real driving logs render every view from the state the model has driven into, and mainly report scores aggregated over whole routes~\cite{hugsim,alpasim,odyssey}.
Others test safety-critical events and find that open-loop strength does not carry over to closed loop.
In Euro NCAP-style collisions with a stopped, an oncoming or a crossing vehicle, UniAD and VAD collide in 89\,\% and 93\,\% of the runs~\cite{neuroncap}.
On 28 kinds of events, the best of 20 learned policies succeeds in 62\,\% of the cases, against 81\,\% for a rule-based planner, and gains in open-loop score do not reliably carry over~\cite{navsafe}.
Such events are harder than everyday driving.

\paragraph{Post-training.}
A policy cloned from human driving compounds its errors once it leaves the states of the driver~\cite{dagger}.
Most post-training of end-to-end driving models still runs on logged states: Hydra-MDP~\cite{hydramdp} and DrivoR~\cite{drivor} distill rule-based trajectory scores into the planner, and AutoVLA~\cite{autovla} and Alpamayo-R1~\cite{alpamayo_r1} use such scores as rewards for reinforcement fine-tuning.
Training on the states that the policy reaches by itself follows two paradigms.
Online reinforcement learning runs on privileged state inputs~\cite{gigaflow,carl} or on camera input in reconstructed scenes~\cite{rad,alpagym} and needs large rollout budgets.
In dataset aggregation (DAgger)~\cite{dagger} and privileged distillation~\cite{lbc,roach}, a teacher labels the states that the student visits.
Recovery data, demonstrations that return from perturbed states, are the other standard remedy for compounding errors~\cite{chauffeurnet,codevilla2019,rap}; our training study uses DAgger and recovery data (Sec.~\ref{sec:remedy}).
Three recent methods are closest to it.
RoaD~\cite{road} guides the rollouts of the policy toward the logged human trajectory, OPTED~\cite{opted} labels the states of a frozen-backbone student with a reinforcement-learning teacher in AlpaSim, and DriveZero~\cite{drivezero} distills such a teacher into a camera policy on logged frames.
None of them had released runnable code and weights at the time of submission, so they are not among the models we test.
\section{LaWanle}
\label{sec:setup}

\subsection{Benchmark}
\label{sec:bench}

LaWanle holds 269 turns extracted from HUGSIM, AlpaSim and Odyssey (Supp.~Sec.~\ref{sup:bench-lawanle}), each with a replay of the recorded human drive (Fig.~\ref{fig:exam}a).
On HUGSIM every turn is expanded into eight copies that differ in the start: a release 25\,m before the junction entry, alone and with a lateral offset of $\pm$0.5\,m, a heading offset of $\pm$3\textdegree{} or a speed offset of $\pm$1\,m/s, and a release 40\,m before the entry, each closer where the scenario begins later (Fig.~\ref{fig:exam}a).
The turns of AlpaSim and Odyssey have no such copies, since on both a start placed at the turn does not reliably reproduce the outcome of the full drive.

We count only the failures of a turn that are strictly attributable to the model, of five kinds: an at-fault collision, leaving the drivable area, not taking the turn, stalling, and not reaching the junction (Fig.~\ref{fig:exam}b).
On every benchmark leaving the drivable area is judged on the footprint of the car.
Crossing a lane marking and running a red light pass (Fig.~\ref{fig:exam}c).
A turn that ends in an event unrelated to turning, such as a collision caused by another vehicle, is removed from the denominator (Fig.~\ref{fig:exam}d).

We run six released models~\cite{uniad,autovla,drivor,garage,wa_jepa,alpamayo_r1} and our policy, each on the inputs closest to its training data and through the client of its authors where one exists (Supp.~Secs.~\ref{sup:serving-models} and~\ref{sup:serving-inputs}).
Our policy is the open-source lane-keeping model of openpilot~\cite{openpilot}, post-trained in closed loop in AlpaSim (Sec.~\ref{sec:remedy}; details in Supp.~Sec.~\ref{sup:remedy-chain}).
\subsection{Results}
\label{sec:results}

\paragraph{Turn pass rates.}
No model passes four turns in five on any benchmark (Table~\ref{tab:exam}), where the recorded human drive passes 88--100\,\% (Supp.~Sec.~\ref{sup:bench-lawanle}); the best is our policy on AlpaSim (75.9\,\%).
Most models turn well on one benchmark and poorly on another: WA-JEPA passes the most turns on HUGSIM (68\,\%) and 9\,\% on AlpaSim, Alpamayo~1.5 passes 66\,\% on AlpaSim and 28\,\% on Odyssey, and UniAD passes 57\,\% on HUGSIM and less than 10\,\% on the other two.
garage varies least, with 54, 41 and 51\,\% on AlpaSim, Odyssey and HUGSIM, and our policy passes the most turns on AlpaSim and Odyssey (56\,\%), where it reaches nearly every junction (NR), and 47\,\% on HUGSIM.
For most models the failures are specific to turns: on Odyssey our policy, garage, WA-JEPA and DrivoR pass 33--56\,\% of the turns and 87--91\,\% of equally long stretches of straight road, while UniAD and AutoVLA struggle even on straight road, where they pass only 54 and 67\,\% of the stretches (Supp.~Sec.~\ref{sup:res-straight}).
NAVSIM does not show the gap: the four released models with a v2 score lie between 88.6 and 91.7 EPDMS, and even on its turn samples alone, under a strict rule, their plans succeed in 86--95\,\% of the cases; in closed loop, under a relaxed rule, the same models pass only 9--68\,\% of the turns (Fig.~\ref{fig:teaser}; Supp.~Sec.~\ref{sup:bench-navsim}).

\paragraph{Home advantage.}
Five models have a benchmark that their authors report on or train on (shaded in Table~\ref{tab:exam}), and each passes its largest share of turns there: 64\,\% on average, against 36\,\% on the other two.
The gap is widest on AlpaSim, where the three models developed on it pass 54--76\,\% of the turns and the other four at most 19\,\%.
HUGSIM separates the models less than the other two benchmarks, as a home benchmark and as an away one: the seven models pass 42--68\,\% of its turns, and UniAD, which was developed on none of the three, passes 57\,\%, beside 54\,\% for DrivoR, which was developed on it.
On Odyssey, the home of no model, no released model passes more than 41\,\%.
WA-JEPA and DrivoR are reported zero-shot on HUGSIM, and garage on AlpaSim: they have not seen its scenes, but may have been selected on its score.

\paragraph{Failure kinds.}
Turns fail in the driving of the car through the junction and seldom through traffic.
Over the six released models, leaving the road ends 38, 39 and 48\,\% of the failed turns on AlpaSim, Odyssey and HUGSIM, and on HUGSIM 95\,\% of these are contacts with the static scene beside the road; an at-fault collision with another vehicle ends 9--10\,\%.
Each model fails in its own way: on HUGSIM, WA-JEPA and DrivoR most often do not take the turn, Alpamayo~1.5 most often stalls, and the other four leave the road (Supp.~Sec.~\ref{sup:models}).
A failure also costs the turns after it: on Odyssey the released models never reach 16--40\,\% of the turns, in seven of ten cases after the benchmark has ended the drive for leaving the route, mostly at an earlier junction (Supp.~Sec.~\ref{sup:res-table}).
In nine of ten failed turns the car hits no other vehicle: what fails is the turn itself, which Sec.~\ref{sec:why} examines.
\section{Why Turns Fail}
\label{sec:why}

\begin{figure*}[t]
\centering
\includegraphics[width=\textwidth]{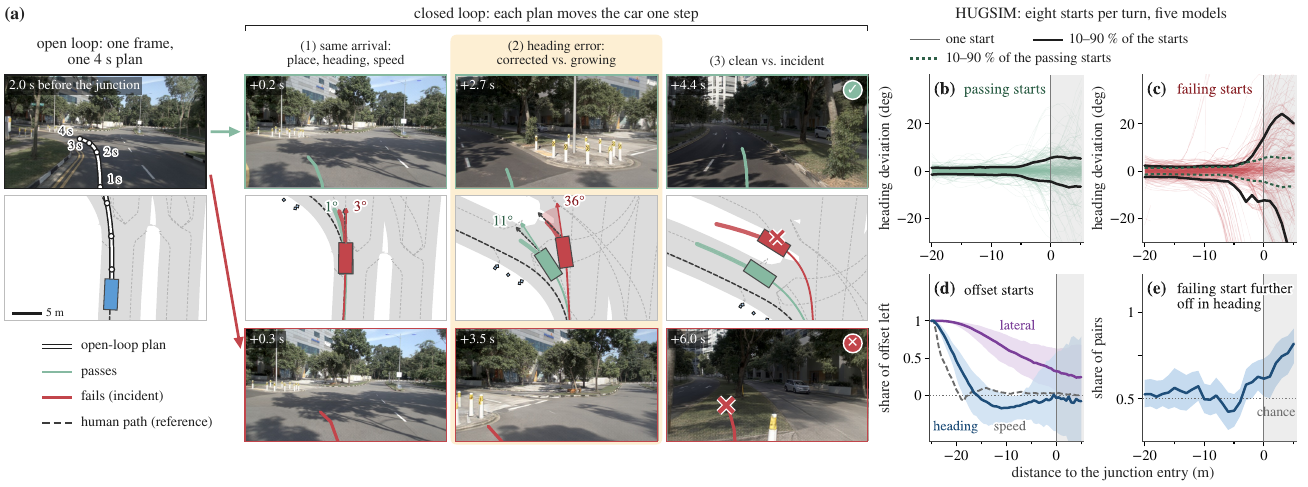}
\caption{\textbf{Passing and failing starts of the same turn.}
(a)~Illustration on one turn: an open-loop plan of our policy (left) and three moments of two of its closed-loop drives (right), with the heading error of each car.
(b),~(c)~Heading of the passing and of the failing starts against the passing starts of the same turn, by distance to the junction entry (shaded: inside the junction; bold: the range of 10--90\,\% of the starts; dashed: that range of the passing starts).
(d)~Share of a start offset that is left (median and interquartile band).
(e)~Share of the failing--passing pairs in which the failing start is further off in heading, with its 95\,\% interval.
(b)--(e): HUGSIM, five models.}
\label{fig:arrival}\label{fig:turnback}
\end{figure*}

A failed turn can have many causes, from vehicle dynamics to perception, and this section does not cover them all.
It studies the turns that a model passes from one start and fails from another.
The usual explanation of a closed-loop failure is compounding error~\cite{dagger}: a cloned policy errs a little, leaves the states of the human driver, and errs more in states its training data do not hold.
From the states of the driver the models do plan the turn (Fig.~\ref{fig:turnback}a, left): at the logged poses of AlpaSim turns, the point 2\,s ahead on the plan of garage leaves a 1\,m corridor around the human path on 2--3\,\% of the frames inside a junction, and on 12\,\% when garage drives on from its own states.
Error accumulates on straight road as well, but there a car that drifts is steered back, and most models complete straight road (Sec.~\ref{sec:results}).
A late turn command~\cite{codevilla2018,navihydra,hua2026command}, an inaccurate route~\cite{hiddenbiases}, a low replan rate and rendering artefacts do not account for these failures either (Supp.~Table~\ref{tab:ruledout}).

Fig.~\ref{fig:turnback}a illustrates this on one turn: two drives of our policy reach the junction with nearly the same position, heading and speed; one corrects its heading error inside the junction, and the other lets it grow and leaves the road.
The rest of this section therefore examines how the selected models recover from an error at the junction.
It first compares the passing and the failing starts of the same turn in the drives of the models themselves, to find where the two begin to differ.
It then intervenes from outside the model: it perturbs the car at the junction to find which error fails a turn that the model otherwise passes, and corrects the plans of a failed turn to find what they lack.

\paragraph{Passing and failing starts.}
\label{sec:census}
On HUGSIM every turn is driven from eight starts (Sec.~\ref{sec:setup}), and on 23--41\,\% of the turns the same model passes from some of them and fails from others (AutoVLA on 69\,\%).
The approach does not decide which.
The offsets given to a start 25\,m before the entry are corrected on the way: those in heading and speed are gone 10\,m before the entry, and a fifth to a half of the lateral offset is left at the entry, depending on the model (Fig.~\ref{fig:arrival}d).
Ten metres before the entry four starts in five head within 2\textdegree{} of the passing starts of the same turn, and two in five of these fail (Fig.~\ref{fig:arrival}b,\,c).

The starts part inside the junction.
Of a failing and a passing start of the same turn, the failing one is further off the human path in heading in 42--63\,\% of the pairs at every distance before the entry, close to an even split, and in 82\,\% of them 5\,m past it (Fig.~\ref{fig:arrival}e).
The difference between a passing and a failing start thus appears at the junction, after most of the error given on the approach has been removed.
For our policy and UniAD the lateral offset at the entry separates a part of the failing starts from the passing ones, and AutoVLA differs from the other models along the whole approach; the numbers per model and the place of the first failure are in Supp.~Sec.~\ref{sup:why-starts}.

\begin{table}[t]
\centering
\footnotesize\setlength{\tabcolsep}{3pt}
\begin{tabular}{@{}lcc@{\hspace{6pt}}ccc@{}}
\toprule
 & \multicolumn{2}{c}{Change of the car} & garage & Ours & WA-JEPA \\
\cmidrule(r{6pt}){2-3}
\emph{(a) Passed turns: failed} & heading & position & {\scriptsize AlpaSim} & {\scriptsize AlpaSim} & {\scriptsize HUGSIM} \\
\midrule
Shift & 0 & 5\,cm & 0\,/\,34 & 1\,/\,54 & 0\,/\,88 \\
Rotation, inside & 2\textdegree & 0 & 1\,/\,17 & 1\,/\,27 & 0\,/\,44 \\
 & 4\textdegree & 0 & 3\,/\,17 & 5\,/\,27 & 0\,/\,44 \\
 & 6\textdegree & 0 & 4\,/\,17 & 6\,/\,27 & 0\,/\,44 \\
Rotation, outside & 4\textdegree & 0 & 0\,/\,17 & 0\,/\,26 & 14\,/\,44 \\
\midrule
 & \multicolumn{2}{c}{Plan corrected in} & garage & Ours & 3 models \\
\cmidrule(r{6pt}){2-3}
\emph{(b) Failed turns: clean} & heading & offset & {\scriptsize AlpaSim} & {\scriptsize AlpaSim} & {\scriptsize Odyssey} \\
\midrule
Heading error & yes & no & 7\,/\,10 & 9\,/\,19 & 20\,/\,42 \\
Heading and offset & yes & yes & 6\,/\,10 & 9\,/\,19 & 17\,/\,42 \\
Control & random & no & 0\,/\,10 & 2\,/\,19 & 1\,/\,42 \\
\bottomrule
\end{tabular}
\caption{\textbf{Perturbing the car and correcting the plan at the junction.}
(a)~Turns that a model passes unperturbed and fails when the car is moved sideways or rotated in place at the junction entry.
(b)~Failed turns that finish without incident when each plan is corrected by the heading error of the car against the human path, by heading error and lateral offset, or, as a control, by the same angle with a random sign.}
\label{tab:interv}
\end{table}

\begin{figure}[t]
\centering
\includegraphics[width=\columnwidth]{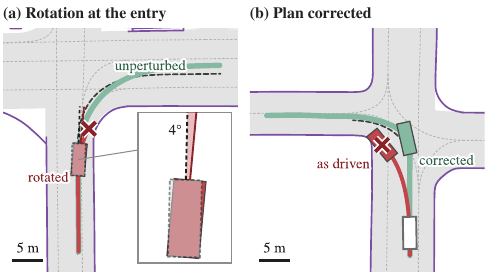}
\caption{\textbf{The two interventions on one turn each} (our policy, AlpaSim).
(a)~The car is rotated by 4\textdegree{} toward the inside at the junction entry (inset: the car before and after) and leaves the road on a turn it passes unperturbed.
(b)~A failed turn passes when each plan is corrected by the heading error of the car.
Cross: the car leaves the road; purple: the road edge; dashed: the human path.}
\label{fig:intervex}
\end{figure}

\paragraph{Interventions.}
\label{sec:mechanism}
The interventions change the pose of the car or the plan handed to the controller from outside the model, and an empty intervention reproduces the stored run exactly.
The first acts on turns that a model passes: 17 AlpaSim turns of garage, 27 of one seed of our policy and the 44 HUGSIM turns that WA-JEPA passes from all eight starts.
Rotating the car in place by 4\textdegree{} at the junction entry changes its heading alone, and it fails 3 of the 17 turns of garage and 5 of the 27 of our policy when it points toward the inside of the turn, and 14 of the 44 of WA-JEPA when it points toward the outside (Table~\ref{tab:interv}a, Fig.~\ref{fig:intervex}a).
Toward the other side it fails no turn of any of the three, and a shift of 5\,cm, the control in position, fails 1 of 176 runs.
A heading error of 4\textdegree{} at the entry thus fails a fifth to a third of the turns that a model otherwise passes, on a side that depends on the model.
A wrong plan held for 0.5\,s inside the junction leaves the car turned by 3--6\textdegree{} and moved by at most 0.1\,m, and it fails turns mostly on the same side (Supp.~Table~\ref{tab:sup-plan}).

The second intervention acts on failed turns: each plan is rotated by the heading error of the car against the human path before the controller receives it, a median 0.7\textdegree{} per call on AlpaSim and 1.8\textdegree{} on Odyssey.
Of 10 failed AlpaSim turns of garage 7 then finish without incident, 9 of 19 of our policy, and 20 of the 42 Odyssey turns that three models fail by running wide; the same angle with a random sign rescues 0, 2 and 1 (Table~\ref{tab:interv}b, Fig.~\ref{fig:intervex}b).
Correcting the lateral offset as well rescues no further turn.
No model has the human path, so the correction is a diagnostic: the plans of these failed turns lack a rotation of about a degree per call in the right direction.
A delay of the fresh plans after a rotation is tested in Supp.~Sec.~\ref{sup:why-interventions}.

On the turns and models tested, these reads locate the failure: a model that plans the turn from the states of the driver and removes an error on the approach meets, at the junction and in a state it reached by its own driving, a heading error of a few degrees and corrects it too little.
What in a model leaves the correction short is not resolved here; Sec.~\ref{sec:remedy} asks whether training on the states of the junction supplies it.
\section{Training for Turns}
\label{sec:remedy}

\begin{figure*}[t]
\centering
\includegraphics[width=\linewidth]{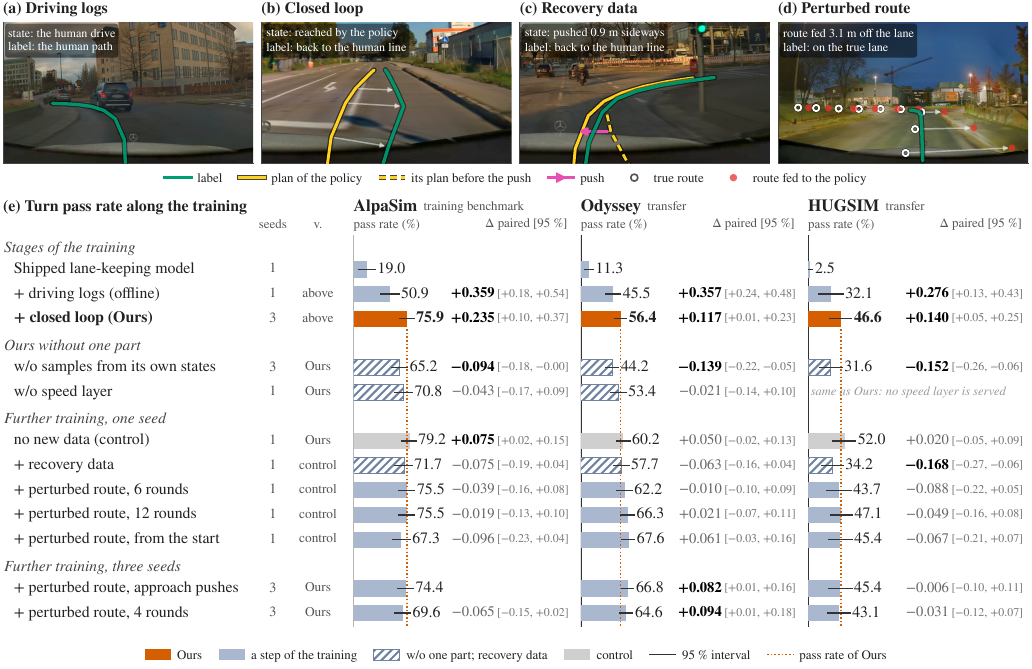}
\caption{\textbf{Training for turns.} (a)--(d): one training sample of each kind of data.
(e): turn pass rate of our policy on the three benchmarks, along its training and under further training.
$\Delta$: paired difference to the row named in `v.', bold when its 95\,\% interval excludes zero.}
\label{fig:training}
\end{figure*}

This section explores how training can make a policy turn, and whether that ability transfers to other benchmarks.
We first describe how our own policy is trained and which data its further training adds, then read what each stage and each part of the training contributes to turns, and last whether further training on states at the junction closes the gap that remains (Fig.~\ref{fig:training}).
`Ours' in every table and figure of the paper is the policy at the end of its closed-loop stage.
Every policy is read on AlpaSim, the benchmark of its closed-loop training, and on Odyssey and HUGSIM, which test the transfer; differences in pass rate are paired over turns and carry a 95\,\% scene-bootstrap interval, and are resolved when it excludes zero (Supp.~Sec.~\ref{sup:remedy}).

\paragraph{Method.}
\label{sec:stages}
Our policy starts from the lane-keeping model of openpilot~\cite{openpilot} (Cinque Terre v3, released in September 2026), which is not designed for turns at junctions.
We freeze its vision encoder and train the layers behind it, 5\,\% of the parameters, in two stages.
The offline stage trains on driving logs: each state is one the human driver was in, and the label is the path the driver took (Fig.~\ref{fig:training}a).
The closed-loop stage follows DAgger~\cite{dagger}: the policy drives in AlpaSim scenes, and each state it reaches is labelled with a lateral path back to the line the human driver kept in the lane (Fig.~\ref{fig:training}b).
The speed remains that of the policy; a rule-based speed layer converts its plan into the speed that is served.
The final training of one seed takes about one GPU-hour on an RTX~6000D, and collecting the closed-loop data about 28 (Supp.~Sec.~\ref{sup:remedy-chain}).
Further training adds two kinds of samples.
Recovery data~\cite{chauffeurnet,codevilla2019} force states a policy seldom reaches: we push the car off its line at junctions in AlpaSim training scenes, label the way back as before, and draw a quarter of each batch from these samples (Fig.~\ref{fig:training}c).
Route perturbation changes the route fed to the policy during this collection: in two of three episodes it is shifted as a whole by up to 3\,m and jittered, while the label still leads back to the true lane (Fig.~\ref{fig:training}d).
A fine-tune continues from a trained seed for 12{,}000 steps, and its control continues the training from the same seed for as many steps with no new data.

\paragraph{Stages of the training.}
\label{sec:ablations}
Each stage adds turns, and the gain transfers.
On the 57 AlpaSim turns the pass rate rises from 19.0\,\% for the shipped model to 50.9\,\% after the offline stage and to 75.9\,\% after the closed-loop stage; on Odyssey and HUGSIM it climbs in the same steps, to 56.4 and 46.6\,\% (Fig.~\ref{fig:training}).
The closed-loop stage adds two things at once, samples labelled from the states of the policy and a rule-based speed layer, and a layer of rules is the first suspect when a benchmark number rises.
We therefore remove each in turn.
Without the speed layer the AlpaSim score falls by 0.122 and the turns stay where they are, on AlpaSim and on Odyssey.
Without the samples labelled from its own states the policy loses between 0.09 and 0.15 of the turns on each of the three benchmarks, every loss resolved.
Serving does not account for the turns either: garage served through our client passes the same turns as with its own (Supp.~Sec.~\ref{sup:serving}).
Of the two, what makes the policy turn, and what transfers, is therefore what it learned from the states of its own driving; the rules on top carry part of the benchmark score.
\paragraph{Further training on junction states.}
\label{sec:recovery}
After the closed-loop stage the policy still fails a quarter of the AlpaSim turns and about half of the turns on the other two benchmarks, and Sec.~\ref{sec:why} locates the failure in a small error at the junction entry that is corrected too little.
The natural next step is to show the policy more of exactly these states: closed-loop collection holds them only as often as the policy drifts there by itself, and recovery data supply them on purpose.
The fine-tune learns what it is shown and drives 23 of 27 AlpaSim turns with an injected push without incident, against 14 for the control.
The lesson does not reach the turns the policy drives by itself: against the control the fine-tune gains on no benchmark and loses 0.168 [0.063, 0.273] of the turns on HUGSIM (Fig.~\ref{fig:training}).
Looking for what it had learned in place of a better turn, we found that its correction leans on the route input.
A route displaced by 1.75\,m moves its plan by 0.38 of the shift, twice as far as that of the control, and Odyssey serves a road-level route that lies metres off the lane at corners.
Of 12 Odyssey turns that Ours drives cleanly, the fine-tune keeps 2 under the served route and 11 when a lane-accurate route replaces it.

\label{sec:routepert}%
If the correction is anchored to the route, a route that cannot be trusted during training should move it to the image, which is the idea of the route perturbation.
It acts as intended: the fine-tune keeps the repair of the push (21 of 27), follows a shifted route no more than the control (0.18 against 0.17), and keeps 6 of the 12 Odyssey turns under the served route.
Two such fine-tunes, each run from all three seeds, then gain on Odyssey over the seeds they start from, by $+0.082$ [$+0.007$, $+0.157$] and $+0.094$ [$+0.014$, $+0.179$], with no gain on HUGSIM or on the AlpaSim turns.
This read as the remedy until the control was trained for as many steps as the fine-tunes.
With no new data it gains $+0.050$ [$-0.022$, $+0.125$] on Odyssey over the seed it starts from, and passes 0.092 [0.020, 0.164] more Odyssey turns than a control trained for half as many steps.
Against this control, which exists for one of the three seeds, none of the three one-seed runs with a perturbed route is resolved on any benchmark; against the shorter control two of them were on Odyssey (Fig.~\ref{fig:training}, Supp.~Sec.~\ref{sup:remedy-arms}).
We therefore do not attribute the Odyssey gain to the route perturbation: most of it is what further steps give a policy with no new data.
The full benchmarks agree: for the first of the two fine-tunes the scores of all three do not change resolvably, and the NAVSIM v2 EPDMS falls by 0.35 [0.13, 0.57].
We also chose the route perturbation after seeing recovery data fail on Odyssey, and under two alternative pass rules the lower end of its gain over the parent seeds touches zero (Supp.~Sec.~\ref{sup:remedy-arms}).
Turning that transfers came from the states the policy reached by its own driving; states it is pushed into teach it to undo the push, and against a control trained as long, no further training on them, with or without a perturbed route, improves turns on Odyssey or HUGSIM.
\section{Limitations}
\label{sec:limitations}

\paragraph{Clients and controllers.}
On its home benchmark a model runs through the client of its authors where one exists, and elsewhere through our adapter, which serves it the inputs closest to its training data; defects of shipped code that we correct are listed with the configuration of every cell (Supp.~Sec.~\ref{sup:serving}).
Each benchmark tracks every plan with one controller, the same for all models on it.
We separated the two parts by swapping each.
On AlpaSim, the client of our policy, its speed layer, is worth 0.12 of its benchmark score and no resolvable share of its turns ($-0.04$ [$-0.17$, $+0.09$]), and garage passes the same turns through the client of our policy as through its own ($+0.04$ [$-0.10$, $+0.19$]; Supp.~Sec.~\ref{sup:serving-swap}).
On Odyssey, a controller of the AlpaSim kind changes on which side the turns of our policy fail and halves the unperturbed turns it passes (Supp.~Sec.~\ref{sup:serving-swap}): the controller is part of what a turn rate measures on a benchmark.
The client swap covers garage and our policy on AlpaSim and the controller swap one model and one seed; the other cells were not swapped.

\paragraph{Size of the HUGSIM part.}
The HUGSIM part of LaWanle holds 99 turns at 22 places in 21 scenes, 15 of the places in KITTI-360, and its intervals resample these scenes.
Pairwise comparisons on it are therefore weak, with 4 of the 21 pairs of models resolved (Supp.~Sec.~\ref{sup:bench-lawanle}), and a pass rate depends on the mix of places: our policy passes 69\,\% of the starts on Waymo scenes and 45\,\% on KITTI-360 scenes.

\paragraph{Overlap with our training data.}
Odyssey is built from nuPlan logs, and 68 of its 100 scenes lie inside drives of the AlpaSim scenes on which our policy was post-trained.
Against the third-party models, whose training logs are disjoint from the benchmark, our policy gains no measurable advantage on these scenes ($+0.03$ [$-0.09$, $+0.16$] in the share of clean turns), a test that excludes only a large advantage.

\paragraph{Mechanism and training.}
The interventions of Sec.~\ref{sec:mechanism} cover three models, and the direction of heading error that fails a turn differs between them, so they give no single mechanism; of the other explanations tested, the entry speed is left open (Supp.~Table~\ref{tab:ruledout}).
The training study fine-tunes one model with a frozen encoder, and the route perturbation was chosen after looking at the benchmarks (Sec.~\ref{sec:remedy}); most of its fine-tunes have one seed and one control run, and the two that were repeated from three seeds gain on Odyssey and not on HUGSIM (Supp.~Sec.~\ref{sup:remedy-arms}).

\paragraph{The human reference.}
The human row is a replay of the recorded drive: it does not react to the traffic that HUGSIM inserts, on Odyssey 12 of its turns leave the mapped drivable area, by less than 0.8\,m, and on AlpaSim it is the log against which the scene was built (Supp.~Sec.~\ref{sup:bench-ceiling}).
In future work, human drivers will drive the turns of HUGSIM, AlpaSim and Odyssey, giving a reference that reacts and is held to the same rule as the models.
\section{Conclusion}
\label{sec:conclusion}
Leading driving models that score close to the human driver on NAVSIM do not reliably make a turn in closed loop, and each model developed on one of the benchmarks turns best there.
The starts of a turn that pass and fail are hard to tell apart on the approach and part inside the junction, where a heading error of a few degrees is corrected too little.
Turning can be trained, as the stages of our policy show.
Further fine-tuning on recovery data improves turns on no benchmark against a control trained equally long, and every variant stays far below the human driver.
We will release LaWanle and recommend that closed-loop results report success per manoeuvre beside the benchmark score.
\section*{LLM Usage Statement}
We used Claude (Anthropic), through Claude Code and chat, to write and debug the experiment code
(benchmark runners, model adapters, evaluation and plotting scripts), to analyse result files,
and to draft and edit text.
The authors designed the study, devised the experiments, formulated the claims, reviewed the code, checked every result
and revised the text, and take full responsibility for the content of the paper.
{
    \small
    \bibliographystyle{ieeenat_fullname}
    \bibliography{references}
}

\clearpage
\onecolumn
\renewcommand{\topfraction}{0.9}\renewcommand{\dbltopfraction}{0.9}\renewcommand{\textfraction}{0.07}
\setcounter{topnumber}{3}\setcounter{dbltopnumber}{3}\setcounter{totalnumber}{5}
\begin{center}\Large\textbf{Appendix: Supplementary Material}\end{center}\vspace{-1.2em}
\appendix
\renewcommand{\thesection}{\Alph{section}}
\renewcommand{\thetable}{S\arabic{table}}
\renewcommand{\thefigure}{S\arabic{figure}}
\setcounter{section}{0}\setcounter{table}{0}\setcounter{figure}{0}
\setcounter{secnumdepth}{2}
\makeatletter\def\lwsubsubsect#1{\lwsubsubsection{#1}}\makeatother
\addtocontents{toc}{\protect\setcounter{tocdepth}{2}}
\begin{figure}[h]
  \centering
  \includegraphics[width=\linewidth]{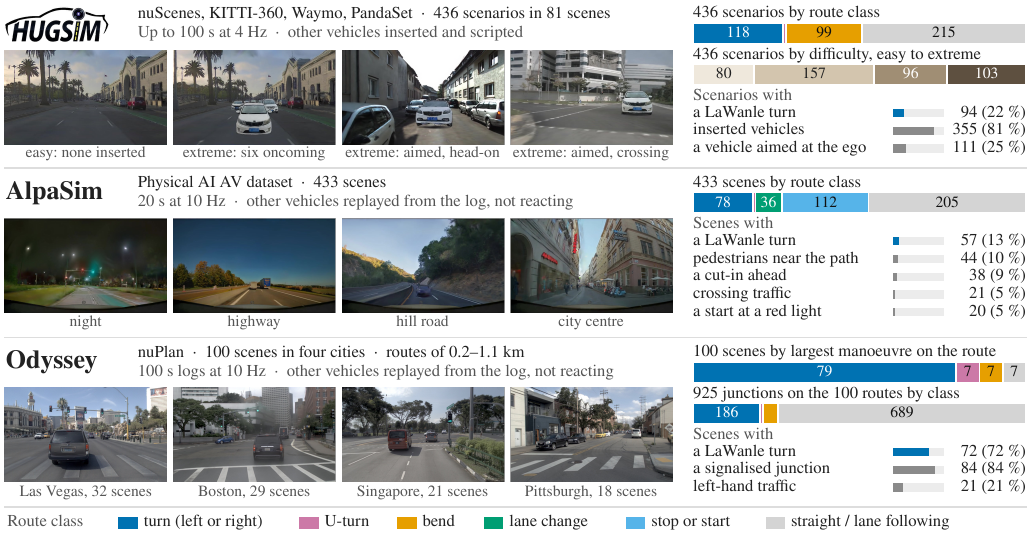}
  \caption{\textbf{The three benchmarks at a glance.}
  Each scene is assigned its largest manoeuvre; HUGSIM difficulty bar (light to dark): easy, medium, hard, extreme.}
  \label{fig:benchmarks}
\end{figure}
{\small\setcounter{tocdepth}{2}\hypersetup{linkcolor=black}\makeatletter\renewcommand{\l@section}[2]{\addpenalty{-\@highpenalty}\vskip 0.35em plus 1pt\begingroup\bfseries\@dottedtocline{1}{0em}{1.5em}{#1}{#2}\endgroup}\makeatother\vspace{-0.8em}\renewcommand{\contentsname}{Contents of the appendix}\tableofcontents}
\twocolumn
\section{Benchmarks and LaWanle}
\label{sup:benchmark}

\subsection{HUGSIM, AlpaSim and Odyssey}
\label{sup:boards}

Fig.~\ref{fig:benchmarks} summarises the three closed-loop benchmarks.

\paragraph{HUGSIM.}
\label{sup:boards-hugsim}
HUGSIM~\cite{hugsim} evaluates 436 scenarios (up to 100\,s at 4\,Hz) across four driving datasets and four difficulty tiers, offering rich visual variety and balanced scenario types.
An episode ends at a collision or more than 10\,m from the recorded route, whereas a step off the drivable area only scores zero and does not end it.
Its harder tiers rely heavily on scripted vehicles, including deliberate attackers rarely seen on real roads, so scores are decided mostly against inserted traffic and many turn-route episodes end in collisions before reaching the junction.
Across garage, WA-JEPA and our policy, failed turns account for only 10--16\,\% of failed episodes and 8--12\,\% of the lost score, leaving the overall score insensitive to turning performance; the recorded human drive is detailed in Sec.~\ref{sup:bench-ceiling} (\protect\nameref{sup:bench-ceiling}).
garage and WA-JEPA thus both score 0.46 on HUGSIM (0.457 and 0.461) but pass 51.3 and 68.1\,\% of its turns (Table~\ref{tab:sup-exam-full}).

\paragraph{AlpaSim.}
AlpaSim~\cite{alpasim} evaluates a rich set of driving capabilities within 433 20\,s clips (10\,Hz) from the Physical AI AV dataset, including lane following, turns, and pedestrian and vehicle interactions, spanning day, dusk and night, varying visibility, highways, hilly roads, urban streets and narrow alleys.
Its logged traffic is non-reactive, so cautious policies are often rear-ended by following vehicles, and the route command can diverge from the logged human path (e.g., signalling a highway lane change where the human stays in lane), penalising models that follow it out of the evaluation corridor.

\paragraph{Odyssey.}
Odyssey~\cite{odyssey} evaluates 100 urban routes (up to 100\,s at 10\,Hz) across four nuPlan~\cite{nuplan} cities on the non-reactive track, with clean rendering even far from the logged trajectory and mostly uncongested streets.
Its coarse polyline route waypoints often lie off-lane, so following them blindly causes collisions (which also blocks route-following shortcuts), and collisions do not end an episode, which favours route completion over strict safety.

\paragraph{Controllers and plan length.}
\label{sup:serving-controller}
Each model reaches the benchmark through its own client, which turns the output of the model into a plan (Sec.~\ref{sup:serving-models}); the controller of the benchmark then tracks that plan and is shared by every model on a benchmark: the model-predictive controller on AlpaSim (weight 10 on changes of the steering angle; AlpaSim ships 5), the two-stage tracker on Odyssey, and on HUGSIM the tracker with its reference heading no longer transposed (Sec.~\ref{sup:serving-models}).
HUGSIM scores a plan over its whole length, so every plan is cut to its first 3\,s before the controller; Table~\ref{tab:reproduction} gives the scores with the shipped plan beside them.
Sec.~\ref{sup:serving-swap} swaps the client and the controller.

\subsection{NAVSIM}
\label{sup:bench-navsim}
\begin{figure}[tbp]
  \centering
  \captionof{table}{\textbf{Turn and straight samples on NAVSIM} (navtest v2; 5\,511 straight, 2\,406 turn samples). $\Delta$: turn minus straight; $\Delta-$human: $\Delta$ minus that of the recorded driver, paired by sample. Right: failure rates on the turn samples, straight samples in brackets; DAC, DDC, LK, EC: drivable area, driving direction, lane keeping, extended comfort. Ours: three seeds.}
  \label{tab:sup-navsim-turns}
  \scriptsize\setlength{\tabcolsep}{1.5pt}
\begin{tabular}{@{}l rr rr@{\hspace{3pt}}|@{\hspace{2pt}}rrrr@{}}
\toprule
 & \multicolumn{4}{c|}{EPDMS} & \multicolumn{4}{c}{Fails on turns (\%)} \\
Model & Str. & Turn & $\Delta$ & $\Delta$ $-$ human & DAC & DDC & LK & EC \\
\midrule
WA-JEPA & 94.4 & \cellcolor{navgood}87.8 & \cellcolor{navgood}\begin{tabular}[t]{@{}r@{}}$-$6.6\\[-2pt]{\tiny{\color{black!62}[$-$7.9,$-$5.4]}}\end{tabular} & \cellcolor{navgood}\begin{tabular}[t]{@{}r@{}}$-$1.9\\[-2pt]{\tiny{\color{black!62}[$-$3.0,$-$0.8]}}\end{tabular} & 3.0\,{\tiny\color{black!62}(0.8)} & \cellcolor{navgood}0.8\,{\tiny\color{black!62}(0.1)} & \cellcolor{navgood}1.8\,{\tiny\color{black!62}(0.7)} & \cellcolor{navgood}32.6\,{\tiny\color{black!62}(2.1)} \\
DrivoR & 92.9 & 84.7 & \begin{tabular}[t]{@{}r@{}}$-$8.2\\[-2pt]{\tiny{\color{black!62}[$-$10.4,$-$6.0]}}\end{tabular} & \begin{tabular}[t]{@{}r@{}}$-$3.4\\[-2pt]{\tiny{\color{black!62}[$-$5.4,$-$1.6]}}\end{tabular} & \cellcolor{navgood}1.7\,{\tiny\color{black!62}(0.6)} & \cellcolor{navbad}5.2\,{\tiny\color{black!62}(0.1)} & \cellcolor{navbad}8.0\,{\tiny\color{black!62}(1.4)} & \cellcolor{navbad}42.9\,{\tiny\color{black!62}(24.3)} \\
AutoVLA & 94.6 & 78.8 & \cellcolor{navbad}\begin{tabular}[t]{@{}r@{}}$-$15.8\\[-2pt]{\tiny{\color{black!62}[$-$17.6,$-$13.8]}}\end{tabular} & \cellcolor{navbad}\begin{tabular}[t]{@{}r@{}}$-$11.0\\[-2pt]{\tiny{\color{black!62}[$-$12.6,$-$9.3]}}\end{tabular} & 9.8\,{\tiny\color{black!62}(1.0)} & 2.5\,{\tiny\color{black!62}(0.1)} & 4.3\,{\tiny\color{black!62}(0.4)} & 42.6\,{\tiny\color{black!62}(2.7)} \\
garage & 91.6 & 82.3 & \begin{tabular}[t]{@{}r@{}}$-$9.3\\[-2pt]{\tiny{\color{black!62}[$-$11.4,$-$7.3]}}\end{tabular} & \begin{tabular}[t]{@{}r@{}}$-$4.6\\[-2pt]{\tiny{\color{black!62}[$-$6.2,$-$3.0]}}\end{tabular} & 5.8\,{\tiny\color{black!62}(1.7)} & 1.5\,{\tiny\color{black!62}(0.2)} & 3.9\,{\tiny\color{black!62}(0.9)} & 41.1\,{\tiny\color{black!62}(7.5)} \\
\ w/o route & 91.6 & 82.0 & \begin{tabular}[t]{@{}r@{}}$-$9.6\\[-2pt]{\tiny{\color{black!62}[$-$11.5,$-$7.6]}}\end{tabular} & \begin{tabular}[t]{@{}r@{}}$-$4.9\\[-2pt]{\tiny{\color{black!62}[$-$6.5,$-$3.2]}}\end{tabular} & 5.3\,{\tiny\color{black!62}(1.6)} & 2.6\,{\tiny\color{black!62}(0.1)} & 4.2\,{\tiny\color{black!62}(1.0)} & 41.3\,{\tiny\color{black!62}(7.0)} \\
Ours & 93.5 & \cellcolor{navbad}78.6 & \begin{tabular}[t]{@{}r@{}}$-$14.9\\[-2pt]{\tiny{\color{black!62}[$-$17.2,$-$12.4]}}\end{tabular} & \begin{tabular}[t]{@{}r@{}}$-$10.2\\[-2pt]{\tiny{\color{black!62}[$-$12.2,$-$8.1]}}\end{tabular} & \cellcolor{navbad}10.3\,{\tiny\color{black!62}(0.8)} & 2.8\,{\tiny\color{black!62}(0.1)} & 4.9\,{\tiny\color{black!62}(0.8)} & \cellcolor{navgood}32.6\,{\tiny\color{black!62}(0.8)} \\
\midrule
\textit{Human} & 96.2 & 91.5 & \begin{tabular}[t]{@{}r@{}}$-$4.7\\[-2pt]{\tiny{\color{black!62}[$-$5.6,$-$3.8]}}\end{tabular} & -- & 0.0\,{\tiny\color{black!62}(0.0)} & 0.4\,{\tiny\color{black!62}(0.0)} & 0.0\,{\tiny\color{black!62}(0.0)} & 29.6\,{\tiny\color{black!62}(0.4)} \\
\bottomrule
\multicolumn{9}{@{}r@{}}{\rule{0pt}{2.4ex}\colorbox{navgood}{best}\ \colorbox{navbad}{worst} of the six models in a column; grey: 95\,\% interval} \\
\end{tabular}

  \vspace{0.9em}
  \includegraphics[width=0.75\linewidth]{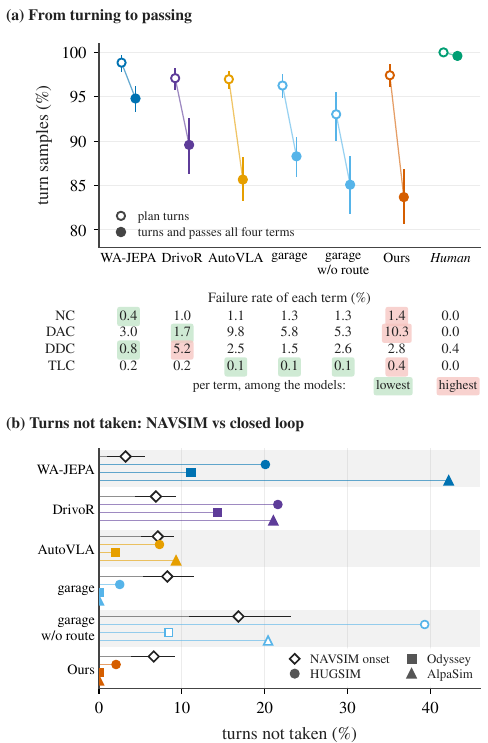}
  \captionof{figure}{\textbf{NAVSIM turn samples against closed-loop turns.} (a)~The 2\,406 turn samples. (b)~Plans that do not turn on the 873 samples taken before the car starts to turn, and closed-loop turns not taken (MT in Table~\ref{tab:exam}). Bars: 95\,\% intervals over logs.}
  \label{fig:sup-navsim-quant}
\end{figure}
\begin{figure}[tbp]
  \centering
  \includegraphics[width=\linewidth]{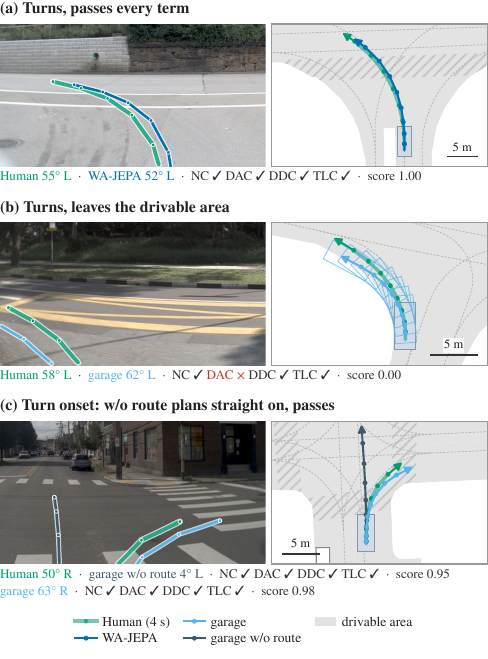}
  \caption{\textbf{Three NAVSIM turn samples.} The 4\,s path of the recorded driver and the plans, projected onto the front camera and drawn over the drivable area of the map; below each row, the headings after 4\,s, the four gates (NC: no at-fault collision, DAC: drivable area, DDC: driving direction, TLC: traffic light) and the score of the sample. (b)~Outlines: the car at the eight planned poses.}
  \label{fig:sup-navsim-examples}
\end{figure}

\paragraph{Turn success.}
The NAVSIM~\cite{navsim} turn success in Fig.~\ref{fig:teaser} is evaluated on the 2\,406 navtest samples where the recorded driver turns by at least 30\textdegree{} within 4\,s.
A plan succeeds if its 4\,s heading reaches at least half of the turn of the driver, in the same direction, and it passes the collision, drivable-area, driving-direction and traffic-light gates of NAVSIM v2; this is stricter than our closed-loop rule (Sec.~\ref{sup:bench-judge} (\protect\nameref{sup:bench-judge})), which lets red-light violations pass.
WA-JEPA, DrivoR, garage and AutoVLA still succeed on 86--95\,\% of the samples and our policy on 84\,\% (Fig.~\ref{fig:sup-navsim-examples}a, Fig.~\ref{fig:sup-navsim-quant}a; UniAD and Alpamayo~1.5 have no NAVSIM read).
Open-loop samples leave little to decide: about half start with the car already mid-turn, and every policy continues the motion (99--100\,\%, even without a route); the command, derived from the logged future, gives the turn direction on 97\,\%; and even at turn onset, road geometry and motion history make 92--97\,\% of plans turn (83\,\% for garage without a route).

\paragraph{What EPDMS penalises on turns.}
Planning straight ahead at turn onset rarely leaves the drivable area within 4\,s and scores almost as high as turning (Fig.~\ref{fig:sup-navsim-examples}c).
EPDMS drops on turns therefore reflect single-step lateral placement rather than missed turns: almost all drivable-area failures occur on plans that do turn (Fig.~\ref{fig:sup-navsim-examples}b; DrivoR loses instead on driving direction and lane keeping), alongside an extended-comfort term (EC) comparing consecutive plans that the recorded human driver fails nearly as often as the models (Table~\ref{tab:sup-navsim-turns}).

\paragraph{Mismatch with closed-loop turning.}
Every NAVSIM plan begins from a logged human state and never executes a full junction under compounding error, so open-loop turn penalties measure a different failure mode from closed-loop turning.
The models that most often drive straight through closed-loop junctions (WA-JEPA, DrivoR and garage without a route) incur smaller open-loop turn penalties than AutoVLA and our policy and still turn reliably at NAVSIM onset, whereas AutoVLA and our policy take the largest EPDMS losses on NAVSIM turns yet almost always initiate the turn in closed loop (Table~\ref{tab:sup-navsim-turns}, Fig.~\ref{fig:sup-navsim-quant}b).
For the same reason, removing the route input barely affects garage on NAVSIM (under 0.5 EPDMS, as straight onset plans remain on road for 4\,s), while costing 7--29 points of closed-loop turn pass rate where choosing and completing the exit is required (Table~\ref{tab:sup-exam-full}).

\subsection{LaWanle}
\label{sup:bench-lawanle}
\begin{figure*}[tbp]
  \centering
  \includegraphics[width=\linewidth]{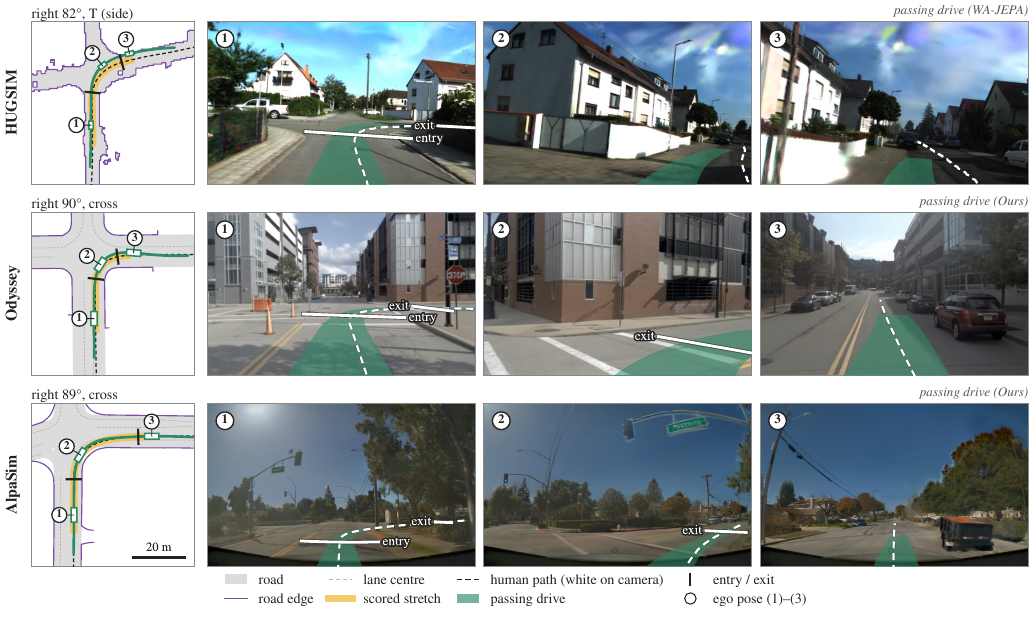}
  \caption{\textbf{One passing turn per benchmark.} Left: the junction from above, at the same scale in every row; the scored stretch runs from 20\,m before the junction entry to 5\,m past the exit, and \textcircled{\scriptsize 1}--\textcircled{\scriptsize 3} mark the poses of the three front-camera frames on the right, on which the paths and the junction entry and exit are projected onto the ground. HUGSIM has no lane map; its road is the reconstructed ground on which leaving the road is judged.}
  \label{fig:sup-turn-gallery}
\end{figure*}

\subsubsection{Turn selection}
\label{sup:bench-turns}
LaWanle tests the most basic turn a driver makes: a left or right turn of about a right angle at a cross or T junction (Fig.~\ref{fig:sup-turn-gallery} shows one per benchmark).
We take a turn only where the human driver of the log drove it, so that its route is defined and the recorded drive serves as a reference, and we take every such turn of each benchmark.
Turns inside which the scene or the route ends are left out, and on Odyssey also turns whose junction entry the car reaches less than 2\,s after the simulator hands it over, so that every turn can be driven and scored in full.
This gives 99 turns on HUGSIM, 57 on AlpaSim and 113 on Odyssey.
The turn lists, the hand labels and a script that fetches only these scenes are in the accompanying archive.

\subsubsection{Varied starts on HUGSIM}
\label{sup:bench-starts}
Driven through the full scenario, a turn is tested only if the car reaches it, and on HUGSIM the episode ends before the car reaches 38--53\,\% of the 99 turns, depending on the model.
We therefore start the car shortly before each HUGSIM turn and drive every turn from eight slightly different starts, which adds episodes to a benchmark of only 22 places and lets passing and failing starts of the same turn be compared (Sec.~\ref{sec:why}, Fig.~\ref{fig:arrival}).
A start at the turn is valid only if it reproduces the outcome of the full drive, and we require it to agree with the full drive at least as often as two perturbed starts of the same turn agree with each other.
It fails this test on the other two benchmarks: on Odyssey it agrees in 9 of 12 cases, against 120 of 120 for two starts, and on AlpaSim in 26 of 30, against 0.92 (garage and one seed of our policy, on the turns with at least 40\,m of approach).
Their turns are therefore scored on the episodes of the full benchmark, in which a model drives each scene once from its beginning; Odyssey episodes are cut 15\,s after the human driver has left the junction.
On HUGSIM it passes, agreeing in 10 of 12 cases (six turns, two checkpoints of our policy) against 0.75 for two starts, and in only 7--9 of 12 without the 2\,s of reference drive described below.
Repeated for every model of Table~\ref{tab:exam}, the nominal start agrees with the full drive in 316 of 363 cases, and the pass rates of the two on the same turns differ by $-$0.10 to $+$0.05 per model, none resolved (pooled $+$0.00 [$-$0.03, $+$0.04]): starting at the turn adds no bias.

The simulator state is set along a reference drive that follows the recorded route at the recorded speed capped at 3\,m/s; the model is fed this drive for 2\,s, with the offset of the start already applied, and the car is released with exactly that offset 25\,m before the junction entry, or 40\,m for the last start (Sec.~\ref{sec:setup}).
Where the scenario begins less than 25\,m before the junction (21 turns at five places), the car is released where it begins, 1.5--16\,m before the entry, without the 2\,s; the 40\,m start lies at 40\,m on 73 of the 99 turns.
The episode ends 5\,m past the junction exit, at a collision, when HUGSIM ends it for leaving the route, after the car has stood for 20\,s, or after 60\,s.

\subsubsection{Pass rule and statistics}
\label{sup:bench-judge}\label{sup:bench-stat}\label{sup:bench-pairs}\label{sup:bench-other}\label{sup:bench-rules}
\begin{table}[tbp]
\centering
\footnotesize\providecolor{kindoffroad}{HTML}{E08E2B}\providecolor{kindcollision}{HTML}{C2454B}\providecolor{kindmissedturn}{HTML}{5B9BD5}\providecolor{kindstall}{HTML}{2E4A7D}\providecolor{kindnotreached}{HTML}{E6E6E6}
\setlength{\tabcolsep}{3pt}
\renewcommand{\arraystretch}{1.08}
\begin{tabular}{@{}>{\raggedright\arraybackslash}p{0.565\linewidth}@{\hspace{4pt}}ccc@{}}
\toprule
Event in the scored stretch & AlpaSim & Odyssey & HUGSIM \\
\midrule
\multicolumn{4}{@{}l}{\textit{Fails the turn}} \\
\hangindent=1.2em\hangafter=1 \textcolor{kindoffroad}{\rule{5pt}{5pt}}\,Leaving the drivable area$^{a}$ & $\times$ & $\times$ & $\times$ \\
\hangindent=1.2em\hangafter=1 \textcolor{kindoffroad}{\rule{5pt}{5pt}}\,Contact with the static scene, parked cars included & {\color{gray}--} & {\color{gray}--} & $\times$ \\
\hangindent=1.2em\hangafter=1 \textcolor{kindcollision}{\rule{5pt}{5pt}}\,At-fault collision with a vehicle & $\times$ & $\times$ & $\times$ \\
\hangindent=1.2em\hangafter=1 \textcolor{kindmissedturn}{\rule{5pt}{5pt}}\,Turning less than a third of the turn, or wrong exit & $\times$ & $\times$ & $\times$ \\
\hangindent=1.2em\hangafter=1 \textcolor{kindstall}{\rule{5pt}{5pt}}\,Stalling within the scored stretch & $\times$ & $\times$ & $\times$ \\
\hangindent=1.2em\hangafter=1 {\setlength{\fboxsep}{0pt}\fcolorbox{gray}{kindnotreached}{\rule{0pt}{4pt}\rule{4pt}{0pt}}}\,Junction entry not reached & $\times$ & $\times$ & $\times$ \\
\midrule
\multicolumn{4}{@{}l}{\textit{Leaves the denominator}} \\
\hangindent=1.2em\hangafter=1 Collision the benchmark assigns to the other party & $\circ$ & $\circ$ & $\circ$ \\
\hangindent=1.2em\hangafter=1 At-fault collision with a standing or slower vehicle ahead, before the entry & $\circ$ & $\circ$ & $\times$ \\
\hangindent=1.2em\hangafter=1 Contact with a pedestrian or cyclist & $\circ$ & $\cdot$ & {\color{gray}--} \\
\hangindent=1.2em\hangafter=1 A replayed vehicle drives into the car outside every oncoming or crossing lane & $\circ$ & $\cdot$ & {\color{gray}--} \\
\hangindent=1.2em\hangafter=1 Release onto a road user, or a collision within 2\,s of release & {\color{gray}--} & {\color{gray}--} & $\circ$ \\
\midrule
\multicolumn{4}{@{}l}{\textit{Passes}} \\
\hangindent=1.2em\hangafter=1 Crossing a lane marking & \checkmark & \checkmark & \checkmark \\
\hangindent=1.2em\hangafter=1 Leaving the route lanes while on the road & \checkmark & \checkmark & \checkmark$^{b}$ \\
\hangindent=1.2em\hangafter=1 Running a red light & \checkmark & \checkmark & \checkmark \\
\midrule
\multicolumn{4}{@{}l@{}}{\scriptsize $\times$ fails\quad $\circ$ leaves the denominator\quad \checkmark\ passes} \\[-1pt]
\multicolumn{4}{@{}l@{}}{\scriptsize $\cdot$ ignored, the turn is scored on what follows\quad {\color{gray}--} no separate case} \\
\bottomrule
\end{tabular}
\caption{\textbf{How an event in the scored stretch is judged.} Coloured squares: the failure kind of Table~\ref{tab:exam}. $^a$A touch shorter than 1\,s included. On AlpaSim and Odyssey the drivable area is that of the map of the benchmark, judged on the footprint of the car; on HUGSIM, which has no map, a pose is off the road when less than half of the 3.0 by 1.6\,m footprint lies over the road points of the reconstructed ground, and off road is not judged at two places (8 turns) whose ground has gaps on the recorded path. $^b$HUGSIM ends an episode when the car is more than 10\,m from the recorded route; the turn then passes only if the car has turned at least a third of the turn, stayed on the drivable area and turned no more than twice the heading the turn needs, and fails off the road otherwise.}
\label{tab:sup-judge}
\end{table}

A turn is scored from 20\,m before the junction entry to 5\,m past the exit.
It fails only on an error of the driving of the model, of one of the five kinds of Table~\ref{tab:exam}, and it leaves the denominator when it ends in an event that is not a turning error, such as a collision the benchmark assigns to the other party or, on the two log-replay benchmarks, a rear-end collision in a queue before the junction, which is a car-following failure (Table~\ref{tab:sup-judge}).
No episode is stopped for being off the road.

\begin{figure}[tbp]
  \centering
  \includegraphics[width=\linewidth]{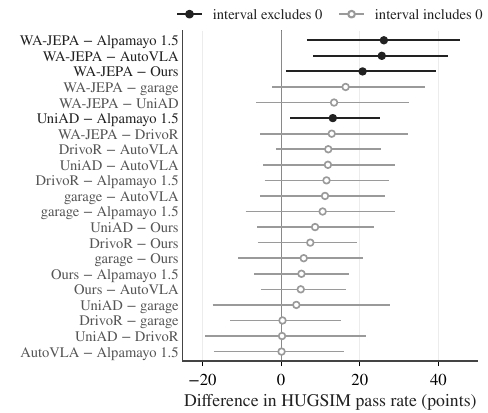}
  \caption{\textbf{Pairs of models on HUGSIM.} Paired difference of the pass rate, first minus second model, on the starts scored for both, with its 95\,\% scene-bootstrap interval; each pair is ordered so that the difference is positive.}
  \label{fig:sup-hug-pairs}
\end{figure}

The pass rate is the share of scored turns that pass, on HUGSIM of scored starts, and its 95\,\% interval comes from a bootstrap over scenes (10\,000 resamples) that draws all turns and starts of a scene together; for our policy the three seeds share each resample and the interval is that of their mean.
Two models are compared by the mean difference per turn on the turns scored for both, under the same bootstrap, and the difference is resolved when its interval excludes zero.
A repeated episode gives the same outcome, so an interval reflects the sample of scenes only.
HUGSIM has the fewest scenes: its 99 turns lie at 22 places in 21 scenes, because the scenarios of a scene share its road and recorded path and differ only in the traffic they insert.
Its intervals are therefore wide, and only 4 of the 21 pairs of models are resolved (Fig.~\ref{fig:sup-hug-pairs}); its rates also depend on the mix of places (Table~\ref{tab:sup-by-dataset}), with five places at which our policy passes fewer than one start in ten and another model at least four in five.

\begin{table}[tbp]
\centering
\footnotesize\setlength{\tabcolsep}{3pt}
\begin{tabular}{@{}l r@{\hspace{3pt}}l r@{\hspace{3pt}}l r@{\hspace{3pt}}l@{}}
\toprule
Model & \multicolumn{2}{c}{KITTI-360 (15)} & \multicolumn{2}{c}{Waymo (5)} & \multicolumn{2}{c}{nuScenes (2)} \\
\midrule
UniAD & 61.3 & {\scriptsize\color{gray}[45,\,77]} & 63.1 & {\scriptsize\color{gray}[20,\,94]} & 16.9 & {\scriptsize\color{gray}[0,\,31]} \\
AutoVLA & 36.6 & {\scriptsize\color{gray}[21,\,49]} & 71.7 & {\scriptsize\color{gray}[47,\,89]} & 9.2 & {\scriptsize\color{gray}[3,\,16]} \\
DrivoR & 49.8 & {\scriptsize\color{gray}[24,\,72]} & \textbf{74.1} & {\scriptsize\color{gray}[48,\,94]} & 36.4 & {\scriptsize\color{gray}[0,\,92]} \\
WA-JEPA & \textbf{71.4} & {\scriptsize\color{gray}[51,\,88]} & 61.4 & {\scriptsize\color{gray}[13,\,95]} & \textbf{60.9} & {\scriptsize\color{gray}[22,\,100]} \\
Alpamayo 1.5 & 42.6 & {\scriptsize\color{gray}[24,\,62]} & 57.7 & {\scriptsize\color{gray}[21,\,86]} & 1.5 & {\scriptsize\color{gray}[0,\,4]} \\
garage & 47.5 & {\scriptsize\color{gray}[23,\,70]} & 68.2 & {\scriptsize\color{gray}[57,\,79]} & 38.5 & {\scriptsize\color{gray}[34,\,43]} \\
\rowcolor{oursbg}Ours & 44.9 & {\scriptsize\color{gray}[26,\,63]} & 69.3 & {\scriptsize\color{gray}[41,\,89]} & 4.1 & {\scriptsize\color{gray}[3,\,5]} \\
\bottomrule
\end{tabular}
\caption{\textbf{HUGSIM pass rate by source dataset} (\%, scored starts, footprint rule), with the 95\,\% scene-bootstrap interval; number of places in brackets; bold, the highest per dataset. With two or five scenes per dataset the intervals of nuScenes and Waymo rest on few clusters.}
\label{tab:sup-by-dataset}
\end{table}

We recomputed Table~\ref{tab:exam} from the stored episodes under looser and stricter pass rules, among them passing every touch of the road edge shorter than 1\,s, failing a HUGSIM pose as soon as a quarter of the footprint is off the road, and counting the contacts removed from the denominator as failures: the best model on each benchmark stays the same, and every model with a home benchmark still passes the most turns there.
Under the stricter HUGSIM test one more pair of models is resolved (DrivoR above AutoVLA) and none is lost.

\subsubsection{Human reference drive}
\label{sup:bench-ceiling}
The last row of Table~\ref{tab:exam} replays the recorded human drive of every turn under the same pass rule.
On HUGSIM it follows the recorded path and speed and does not yield to the traffic the scenario inserts: all its failed turns are collisions with this traffic, and replayed through the full scenarios it scores 0.88 on the scenarios without inserted vehicles and 0.15 on those with them.
On Odyssey its failures are corners of the footprint at most 0.78\,m past the drivable area of the scorer, which follows the nuPlan lane surfaces and ends 1--6\,m inside the kerb; none is a contact.
On AlpaSim the scene is reconstructed from the same log, and the drive passes every turn.
The replay is thus no closed-loop human baseline; a full one, with several human drivers driving the turns of all three benchmarks in closed loop under the same rule as the models, is future work (Sec.~\ref{sec:limitations}).
\section{Models and Serving}
\label{sup:serving}

\subsection{Models}
\label{sup:serving-models}
Table~\ref{tab:sup-models} lists the models of Table~\ref{tab:exam} and the client through which each runs on each benchmark, which is the configuration of each cell.
A client written by the authors of a model is used where one exists; every other cell runs through our adapter around the released inference code of the model.

\begin{table*}[tbp]
\centering
\footnotesize\setlength{\tabcolsep}{4pt}
\begin{tabular}{@{}l>{\raggedright\arraybackslash}p{0.17\linewidth}>{\raggedright\arraybackslash}p{0.25\linewidth}>{\raggedright\arraybackslash}p{0.11\linewidth}lll@{}}
\toprule
 & & & & \multicolumn{3}{c}{Client} \\ \cmidrule(l){5-7}
Model & Paradigm & Checkpoint (training data) & Navigation & HUGSIM & AlpaSim & Odyssey \\
\midrule
UniAD~\cite{uniad} & modular & uniad\_base\_e2e (nuScenes) & command & HUGSIM authors & ours & ours \\
AutoVLA~\cite{autovla} & vision-language-action & PDMS 89, Qwen2.5-VL-3B (NAVSIM) & command as text & ours & ours & ours \\
DrivoR~\cite{drivor} & trajectory scorer & Nav1, 25 epochs (NAVSIM) & command & ours & authors$^a$ & ours \\
WA-JEPA~\cite{wa_jepa} & world-model planner & release (nuPlan video, NAVSIM) & command & authors & ours & ours \\
Alpamayo 1.5~\cite{alpamayo_r1} & vision-language-action, reasoning, 10\,B & Alpamayo-1.5-10B & route as text & ours & authors$^b$ & ours \\
garage~\cite{garage} & TransFuser-style regression~\cite{transfuser} & v0.2.0 LTF-base, post-training, seed 0 (nuPlan, PhysicalAI-AV) & target points on the route & ours & authors & ours \\
\midrule
Ours & lane keeping, post-trained in closed loop & three seeds$^c$ & route and command & ours & authors$^c$ & ours \\
\bottomrule
\end{tabular}
\caption{\textbf{Models and clients.}
Navigation: the form each model was trained with.
Client: authors, written by the authors of the model; HUGSIM authors, the UniAD client written by the authors of HUGSIM; ours, our adapter.
$^a$The AlpaSim driver published by the group of the authors of DrivoR, which we feed the command 20\,m ahead on the recorded path, as in its NAVSIM training, in place of 42\,m.
$^b$The driver that ships with AlpaSim, with the navigation text built from the merged route and each call seeded.
$^c$The serving our policy was post-trained with; its row in Table~\ref{tab:exam} is the mean of three seeds.}
\label{tab:sup-models}
\end{table*}

\paragraph{Defects of shipped code that we correct.}
The defects we correct in the benchmarks apply to every model on the benchmark: the HUGSIM controller transposed the reference heading it tracks, the HUGSIM tracker stopped each solve after 0.05\,s of wall-clock time, so that an episode did not repeat under load, the AlpaSim scorer measured the distance to the end of the trajectory wrongly, and on Odyssey a linear-algebra library miscomputed on our processor; the first and the third are also corrected upstream.
Two clients are corrected for one model each.
The HUGSIM client of UniAD normalised its frames as BGR, disabled the ego-motion alignment of the previous BEV, applied the plan 0.73\,m ahead of its origin and read the heading from an Euler angle that folds past $\pm$90\textdegree; none of these corrections moves the score of UniAD beyond noise (Table~\ref{tab:reproduction}).
The AlpaSim driver of Alpamayo~1.5 decodes images on the GPU, and the model then returns non-finite plans on our machine; we decode them on the CPU.
The model code and weights are unchanged throughout, and the configuration of every cell, our adapters included, is released with the code.

\paragraph{Reproduction of published scores.}
\label{sup:serving-repro}
A published score counts as reproduced when it lies inside our 95\,\% interval; no client was tuned to close a gap (Table~\ref{tab:reproduction}).

\begin{table*}[tbp]
\centering
\scriptsize\setlength{\tabcolsep}{3pt}
\begin{tabular}{@{}>{\raggedright\arraybackslash}p{0.1\linewidth}>{\raggedright\arraybackslash}p{0.15\linewidth}>{\raggedright\arraybackslash}p{0.06\linewidth}>{\raggedright\arraybackslash}p{0.072\linewidth}>{\raggedright\arraybackslash}p{0.1\linewidth}>{\raggedright\arraybackslash}p{0.05\linewidth}>{\raggedright\arraybackslash}p{0.1\linewidth}>{\raggedright\arraybackslash}p{0.125\linewidth}>{\raggedright\arraybackslash}p{0.115\linewidth}@{}}
\toprule
 & & & & & & & \multicolumn{2}{c@{}}{From shipped to served, 95\,\% interval} \\ \cmidrule(l){8-9}
Model & Benchmark, metric, scenes & Client as shipped & Published & Ours, client as shipped & Repro\-duced & Ours, as served & Change & Paired difference \\
\midrule
\multicolumn{9}{@{}l}{\textit{Closed loop}} \\
\addlinespace[2pt]
WA-JEPA~\cite{wa_jepa} & HUGSIM HD-Score, 436 & authors & 0.4462 & 0.445\newline[0.412, 0.480] & yes & 0.461\newline[0.426, 0.494] & plan 4 s $\rightarrow$ 3 s & +0.013\newline[$-$0.010, +0.035] \\
 &  &  &  &  &  &  & training cameras & +0.002\newline[$-$0.025, +0.028] \\
\dashmidrule
UniAD~\cite{uniad} & HUGSIM HD-Score, 436 & HUGSIM & 0.3124$^{a}$ & 0.310\newline[0.282, 0.337] & yes$^{a}$ & 0.327\newline[0.296, 0.359] & client corrections & +0.010\newline[$-$0.010, +0.029] \\
 &  &  &  &  &  &  & training cameras & +0.007\newline[$-$0.021, +0.035] \\
\dashmidrule
DrivoR~\cite{drivor} & HUGSIM HD-Score, 436 & ours & 0.357$^{b}$ & 0.316\newline[0.284, 0.349] & \textbf{no} & 0.322\newline[0.290, 0.355] & plan 4 s $\rightarrow$ 3 s & \textbf{+0.037}\newline\textbf{[+0.026, +0.050]} \\
 &  &  &  &  &  &  & training cameras & \textbf{$-$0.032}\newline\textbf{[$-$0.052, $-$0.011]} \\
\dashmidrule
DrivoR~\cite{drivor} & Odyssey RouteDS, 100 & Odyssey & 20.6$^{c}$ & 21.46\newline[15.94, 27.29] & yes & 24.39\newline[18.97, 30.21] & checkpoint of the authors, training lens & +2.93\newline[$-$2.08, +7.94] \\
\dashmidrule
garage~\cite{garage} & AlpaSim score, 433 & authors & 76.0$^{d}$ & 0.774\newline[0.737, 0.809] & yes$^{d}$ & as shipped & -- & -- \\
\dashmidrule
garage w/o route & AlpaSim score, 433 & authors & 63.9 $\pm$ 1.5$^{d}$ & 0.668$^{e}$ & -- & as shipped & -- & -- \\
\dashmidrule
Alpamayo 1.5~\cite{alpamayo_r1} & AlpaSim score, 433 & authors & none$^{f}$ & failed$^{g}$ & -- & 0.610\newline[0.566, 0.653] & images decoded on the CPU & -- \\
\dashmidrule
Alpamayo 1.5~\cite{alpamayo_r1} & Odyssey RouteDS, 100 & ours & -- & -- & -- & 19.4\newline[13.5, 25.9] & -- & -- \\
\dashmidrule
Alpamayo 1.5~\cite{alpamayo_r1} & HUGSIM HD-Score, 436 & ours & -- & -- & -- & 0.329\newline[0.294, 0.364] & -- & -- \\
\midrule
\multicolumn{9}{@{}l}{\textit{Open loop (the only published score of AutoVLA)}} \\
\addlinespace[2pt]
AutoVLA~\cite{autovla} & NAVSIM PDMS, navtest & -- & 89.11 & 89.53\newline[88.47, 90.52] & yes & -- & -- & -- \\
\addlinespace[2pt]
DrivoR~\cite{drivor} & NAVSIM PDMS, navtest & -- & 93.7 & 93.69\newline[93.12, 94.24] & yes & -- & -- & -- \\
\addlinespace[2pt]
WA-JEPA~\cite{wa_jepa} & NAVSIM PDMS, navtest & -- & 91.8 & 91.70\newline[90.94, 92.43] & yes & -- & -- & -- \\
\addlinespace[2pt]
garage~\cite{garage} & NAVSIM PDMS, navtest & -- & -- & 88.55\newline[87.46, 89.59] & -- & -- & -- & -- \\
\bottomrule
\end{tabular}
\caption{\textbf{Reproduction of published scores.}
Client: HUGSIM, written by the authors of HUGSIM; ours, none released; Odyssey, the baseline shipped with it (non-reactive). Reproduced: the published score lies inside our 95\,\% scene-bootstrap interval. As served: training-matched inputs and, on HUGSIM, a 3\,s plan, each change a separate paired read. Bold: not reproduced, or an interval that excludes zero. AlpaSim: fractions ours, percent published.
$^a$Re-run by the authors of WA-JEPA; none by the authors of UniAD.
$^b$345 scenarios, evaluator not public; the authors of WA-JEPA report 0.325 on the 436.
$^c$For the DrivoR baseline of Odyssey; as served, the checkpoint of the authors of DrivoR with its training lens.
$^d$441 scenes, mean of three seeds; paired with the per-scene results of the authors, $+$0.014 [$-$0.009, $+$0.037].
$^e$No interval read.
$^f$The model card gives 1.37\,$\pm$\,0.10 on 913 scenes in another metric; OPTED~\cite{opted} reports 55.2 on 441.
$^g$Non-finite plans on our machine (Sec.~\ref{sup:serving-models}).}
\label{tab:reproduction}
\end{table*}

\subsection{Serving}
\label{sup:serving-inputs}
\paragraph{Inputs closest to the training data.}
Each model runs with its released code and weights and is served the inputs closest to its training data: its camera rig at the training height and lens, its frame spacing, its ego-state fields and its form of navigation, taken from the documentation and calibration files of its training data and fixed before any read (Table~\ref{tab:sup-serving}).
HUGSIM renders extra cameras that only the model sees, at the training pose of each model, while the scored vehicle, the other road users and the recorded video stay unchanged.
AlpaSim and Odyssey render fixed camera sets, so the views of a model trained on another rig are synthesised there by rotating the rays of the rendered cameras, which can sit up to about 2\,m from the training ones.

\begin{table*}[tbp]
\centering
\scriptsize\setlength{\tabcolsep}{3.5pt}
\begin{tabular*}{\linewidth}{@{\extracolsep{\fill}}llllllllll@{}}
\toprule
 & \multicolumn{3}{c}{HUGSIM, 4\,Hz} & \multicolumn{3}{c}{AlpaSim, 10\,Hz} & \multicolumn{3}{c}{Odyssey, 10\,Hz} \\ \cmidrule(l){2-4} \cmidrule(l){5-7} \cmidrule(l){8-10}
Model & Cameras & Calls & Navigation & Cameras & Calls & Navigation & Cameras & Calls & Navigation \\
\midrule
UniAD~\cite{uniad} & N, 6$^{r}$ & 2\,Hz & cmd & R, 6, rear 28\,\% & 2\,Hz & cmd & R, 6, $\le$14\,\%, rear 62\,\% & 2\,Hz & cmd \\
AutoVLA~\cite{autovla} & N, 3 & each & cmd text & R, 3, $\le$5\,\% & 2\,Hz & cmd text & N, 3, 19\,\% & 2\,Hz & cmd text \\
DrivoR~\cite{drivor} & N, 4$^{r}$ & each & cmd & R, 4 & each & cmd & N, 4, 19\,\% & each & cmd \\
WA-JEPA~\cite{wa_jepa} & N, 4$^{r}$ & each & cmd & R, 4, $\le$6\,\% & 2\,Hz & cmd & N, 4, 19\,\% & each & cmd \\
Alpamayo 1.5~\cite{alpamayo_r1} & N, 4 & each & route text & N, 4 & each & route text & R, 4, wide 40\,\% & 2\,Hz & route text \\
garage~\cite{garage} & N, 1 & cache & targets & N, 1 & cache & targets & N, 1, 19\,\% & cache & targets \\
garage w/o route & N, 1 & cache & none & N, 1 & cache & none & N, 1, 19\,\% & cache & none \\
\midrule
Ours & N, 1 & each & cmd, desire & N, 1 & each & route, cmd & N, 1 & each & route, cmd \\
\bottomrule
\end{tabular*}
\caption{\textbf{Serving of each model on each benchmark}, with the step rate of each.
Cameras: N, the training cameras rendered at their training pose and lens (on HUGSIM, extra cameras only the model sees); R, synthesised by rotating the rays of the cameras of the benchmark; then the number of cameras and the largest share of a view left black (not measured for DrivoR on AlpaSim). $^r$Rear cameras black on Waymo and KITTI-360.
Calls: each, every step; 2\,Hz, plan held between calls; cache, every step with a new plan every 0.5\,s.
Navigation: cmd, the three-way command; targets, points 40, 60 and 80\,m along the route; desire, the command also fed as the turn-desire input of openpilot.
AlpaSim sends no command and HUGSIM no route; both are taken from the recorded path of the scene.}
\label{tab:sup-serving}
\end{table*}

\paragraph{Mismatches that remain.}
UniAD runs with its ego-state network switched off on all three benchmarks, as the HUGSIM client of its authors configures it; with the network on its AlpaSim score is 0.213 instead of 0.119 ($+$0.094 [$+$0.061, $+$0.127]), and on HUGSIM it stalls in 11 of 24 episodes.
The command of a benchmark is defined differently from that of the training data: the HUGSIM command changes less than 5\,m before the junction entry at 7 of 19 places, and the Odyssey command is fed a median 11\,m before the entry; an earlier command rescues few failed turns (Sec.~\ref{sup:why}).

\paragraph{Our policy.}
Our policy has a rule-based speed layer that keeps the path of its plan and sets the speed along it; the layer runs in full on AlpaSim, without its route-hold and takeover rules on Odyssey, and not at all on HUGSIM, where two plan rules of our adapter (backward segments zeroed, a plan ending within 1\,m straightened) change the plan on about 1--2\,\% of the steps.
HUGSIM serves no route, so our adapter feeds the command of the benchmark both as a one-hot and as the turn-desire input of the underlying openpilot model~\cite{openpilot}, which our policy was post-trained with at zero; switching the desire off halves the turns passed, from 25 of 53 to 13 of 51 scored turns of 58 re-run, and the car keeps straight on 12 instead of 4; the HUGSIM cells of our policy are read with it on.
\subsection{Sensitivity to client and controller}
\label{sup:serving-swap}
We swapped the client on AlpaSim for garage and our policy, and the controller on Odyssey for one seed of our policy (Table~\ref{tab:sup-swap}).
The speed layer of our policy carries a tenth of its AlpaSim score and no resolvable share of its turns; served through that layer, garage keeps its score and its turns.
The post-training gain of our policy remains without the layer.
A controller of the AlpaSim kind, which follows the plan between 1 and 2\,s where the Odyssey controller follows its first second, carries out more of each correction: it rescues turns rotated outward and fails turns as driven by cutting the corner.
The side on which turns fail thus depends on the controller of the benchmark, which is the same for every model on it (Sec.~\ref{sup:serving-controller}).

\begin{table*}[tbp]
\centering
\footnotesize\setlength{\tabcolsep}{3pt}
\begin{tabular}{@{}lllllll@{}}
\toprule
\multicolumn{7}{@{}l}{\textit{AlpaSim: the speed layer of our policy}} \\
Model & Serving & Against & Score & $\Delta$ score [95\,\%] & Turns passed ($n$) & $\Delta$ turns [95\,\%] \\
\midrule
Ours & speed layer, seed 0 & -- & 0.908 & -- & 0.750 (52) & -- \\
 & raw plan & speed layer & 0.785 & \textbf{$-$0.122 [$-$0.157, $-$0.088]} & 0.708 (48) & $-$0.043 [$-$0.174, $+$0.087] \\
 & offline stage, raw plan & raw plan, final & 0.466 & \textbf{$-$0.319 [$-$0.369, $-$0.271]} & 0.340 (47) & \textbf{$-$0.349 [$-$0.535, $-$0.163]} \\
\addlinespace[2pt]
garage & own client & -- & 0.774 & -- & 0.542 (48) & -- \\
 & our speed layer, its set speed & own client & 0.756 & $-$0.018 [$-$0.056, $+$0.020] & 0.604 (53) & $+$0.042 [$-$0.104, $+$0.188] \\
 & our speed layer, speed of its plan & own client & 0.770 & $-$0.004 [$-$0.022, $+$0.014] & 0.540 (50) & 0.000 [0.000, 0.000] \\
\midrule
\multicolumn{7}{@{}l}{\textit{Odyssey: the tracking controller, one seed of our policy}} \\
\multicolumn{3}{@{}l}{} & Odyssey & AlpaSim kind & Placebo & \\
\cmidrule(lr){4-6}
\multicolumn{3}{@{}l}{Turns as driven, clean} & \textbf{12 / 12} & 6 / 12 & -- & \\
\multicolumn{3}{@{}l}{Rotated 4\textdegree{} outward, clean} & 5 / 12 & \textbf{9 / 12} & 6 / 12 & \\
\multicolumn{3}{@{}l}{Rotated 4\textdegree{} inward, clean} & 7 / 12 & 4 / 12 & 4 / 12 & \\
\multicolumn{3}{@{}l}{Too wide, made clean} & 0 / 14 & 4 / 14 & 0 / 14 & \\
\multicolumn{3}{@{}l}{Planned correction carried out per 0.5 s} & 0.083 [0.066, 0.100] & \textbf{0.22 [0.18, 0.27]} & -- & \\
\bottomrule
\end{tabular}
\caption{\textbf{Swapping the client and the controller.}
Top: 433 scenes and the 57 AlpaSim turns of Table~\ref{tab:exam}, one read per row; raw plan, the plan without the speed layer; $\Delta$, paired against the row in Against, bold when its interval excludes zero.
Bottom: rotated, the car turned by 4\textdegree{} in the turn; too wide, failed on the outer side under the Odyssey controller; placebo, an edit of the same size with a random sign every 0.5\,s; clean, passed under the rule of Table~\ref{tab:exam}; bold, the clearly better controller (a third more turns, or a higher interval).}
\label{tab:sup-swap}
\end{table*}
\section{Turn Results in Full}
\label{sup:results}

\subsection{The numbers of Table~\ref{tab:exam} in full}
\label{sup:res-table}
Table~\ref{tab:sup-exam-full} gives every cell of Table~\ref{tab:exam} with its scored number of turns, the pass rate on the turns the model reached, the three seeds of our policy, garage without its route (Sec.~\ref{sup:garage-noroute}) and the lane-keeping model from which our policy is post-trained.
On AlpaSim the rule exempts collisions caused by replayed vehicles and contacts while following a car, which removes up to a third of the turns of a model.

\begin{table*}[tbp]
\centering
\footnotesize\setlength{\tabcolsep}{4.2pt}
\begin{tabular}{@{}l r r@{\hspace{4pt}}l rrrrrr r@{}}
\toprule
Model & $n$ & \multicolumn{2}{c}{Pass [95\,\%]} & OR & {\scriptsize of which static} & CO & MT & ST & NR & Reached ($n$) \\
\midrule
\multicolumn{11}{@{}l}{\textit{AlpaSim (57 turns)}} \\
UniAD & 46 & 4.3 & {\color{gray}[0,\,11]} & 50 &  & 9 & 2 & 0 & 35 & 6.7 (30) \\
AutoVLA & 43 & 18.6 & {\color{gray}[7,\,30]} & 40 &  & 7 & 9 & 9 & 16 & 22.2 (36) \\
DrivoR & 38 & 13.2 & {\color{gray}[3,\,24]} & 18 &  & 18 & 21 & 13 & 16 & 15.6 (32) \\
WA-JEPA & 45 & 8.9 & {\color{gray}[2,\,18]} & 31 &  & 2 & 42 & 4 & 11 & 10.0 (40) \\
Alpamayo 1.5 & 50 & \cellcolor{homebg}66.0 & \cellcolor{homebg}{\color{gray}[52,\,78]} & 8 &  & 2 & 6 & 16 & 2 & 67.3 (49) \\
garage & 48 & \cellcolor{homebg}54.2 & \cellcolor{homebg}{\color{gray}[40,\,69]} & 17 &  & 4 & 0 & 17 & 8 & 59.1 (44) \\
\quad w/o route & 49 & 46.9 & {\color{gray}[33,\,61]} & 6 &  & 4 & 20 & 14 & 8 & 51.1 (45) \\
\rowcolor{oursbg}Ours & 52.3 & \cellcolor{homebg}\textbf{75.9} & \cellcolor{homebg}{\color{gray}[65,\,86]} & 7 &  & 10 & 0 & 6 & 1 & 76.4 (52) \\
\quad seeds 0 / 1 / 2 & 52 / 54 / 51 & \multicolumn{9}{l}{75.0 / 72.2 / 80.4} \\
openpilot, as shipped$^\ddagger$ & 42 & 19.0 & {\color{gray}[7,\,31]} & 48 &  & 10 & 2 & 5 & 17 &  \\
\cmidrule(l){1-11}
\textit{Recorded human drive}$^\dagger$ & 57 & 100.0 & {\color{gray}[94,\,100]} & 0 &  & 0 & 0 & 0 & 0 &  \\
\midrule
\multicolumn{11}{@{}l}{\textit{Odyssey (113 turns)}} \\
UniAD & 109 & 9.2 & {\color{gray}[4,\,15]} & 52 &  & 4 & 3 & 0 & 32 & 13.5 (74) \\
AutoVLA & 101 & 12.9 & {\color{gray}[6,\,20]} & 38 &  & 5 & 2 & 3 & 40 & 21.3 (61) \\
DrivoR & 98 & 37.8 & {\color{gray}[27,\,49]} & 19 &  & 11 & 14 & 1 & 16 & 45.1 (82) \\
WA-JEPA & 108 & 33.3 & {\color{gray}[24,\,43]} & 19 &  & 10 & 11 & 5 & 22 & 42.9 (84) \\
Alpamayo 1.5 & 103 & 28.2 & {\color{gray}[19,\,38]} & 17 &  & 10 & 16 & 1 & 28 & 39.2 (74) \\
garage & 106 & 40.6 & {\color{gray}[32,\,50]} & 26 &  & 5 & 0 & 12 & 16 & 48.3 (89) \\
\quad w/o route & 107 & 21.5 & {\color{gray}[13,\,31]} & 15 &  & 8 & 8 & 7 & 40 & 35.9 (64) \\
\rowcolor{oursbg}Ours & 101.7 & \textbf{56.4} & {\color{gray}[47,\,65]} & 20 &  & 22 & 0 & 1 & 1 & 56.8 (101) \\
\quad seeds 0 / 1 / 2 & 101 / 103 / 101 & \multicolumn{9}{l}{56.4 / 57.3 / 55.4} \\
openpilot, as shipped$^\ddagger$ & 106 & 11.3 & {\color{gray}[6,\,18]} & 20 &  & 9 & 1 & 3 & 56 &  \\
\cmidrule(l){1-11}
\textit{Recorded human drive}$^\dagger$ & 113 & 89.4 & {\color{gray}[83,\,95]} & 11 &  & 0 & 0 & 0 & 0 &  \\
\midrule
\multicolumn{11}{@{}l}{\textit{HUGSIM (99 turns $\times$ 8 starts = 792 cells)}} \\
UniAD & 688 & 57.1 & {\color{gray}[42,\,72]} & 36 & 33 & 5 & 0 & 0 & 1 & 57.9 (679) \\
AutoVLA & 671 & 41.9 & {\color{gray}[28,\,55]} & 43 & 40 & 6 & 7 & 0 & 2 & 42.6 (659) \\
DrivoR & 690 & \cellcolor{homebg}54.3 & \cellcolor{homebg}{\color{gray}[35,\,72]} & 14 & 14 & 6 & 22 & 1 & 3 & 56.0 (670) \\
WA-JEPA & 677 & \cellcolor{homebg}\textbf{68.1} & \cellcolor{homebg}{\color{gray}[51,\,83]} & 6 & 6 & 3 & 20 & 0 & 2 & 69.7 (661) \\
Alpamayo 1.5 & 666 & 41.9 & {\color{gray}[25,\,58]} & 17 & 17 & 2 & 6 & 30 & 3 & 43.2 (646) \\
garage & 678 & 51.3 & {\color{gray}[34,\,67]} & 22 & 21 & 5 & 3 & 17 & 3 & 52.9 (658) \\
\quad w/o route & 679 & 22.7 & {\color{gray}[8,\,38]} & 15 & 14 & 4 & 39 & 16 & 4 & 23.5 (654) \\
\rowcolor{oursbg}Ours & 679.7 & 46.6 & {\color{gray}[31,\,62]} & 45 & 35 & 5 & 2 & 0 & 1 & 47.3 (670) \\
\quad seeds 0 / 1 / 2 & 682 / 685 / 672 & \multicolumn{9}{l}{44.1 / 51.4 / 44.3} \\
openpilot, as shipped$^\ddagger$ & 677 & 2.5 & {\color{gray}[0,\,7]} & 46 & 31 & 4 & 2 & 34 & 12 & 2.8 (598) \\
\cmidrule(l){1-11}
\textit{Recorded human drive}$^\dagger$ & 83 & 88.0 & {\color{gray}[80,\,95]} & 0 &  & 12 & 0 & 0 & 0 &  \\
\bottomrule
\end{tabular}
\caption{\textbf{Table~\ref{tab:exam} in full.} $n$: scored turns (HUGSIM: scored starts; Ours: mean over three seeds). Pass rate (\%) with its 95\,\% scene-bootstrap interval; bold, the highest on a benchmark; {\setlength{\fboxsep}{0.8pt}\colorbox{homebg}{shaded}}, the home benchmark of the model, as in Table~\ref{tab:exam}. Share of $n$ (\%) lost to each kind of failure: OR off the drivable area, judged on the footprint of the car, on HUGSIM with contacts with the static scene (more than 100 background points inside the box of the car; parked cars count, the sidewalk does not), listed apart as `of which static'; CO at-fault collision; MT turn not taken (straight on or the wrong way); ST stall; NR junction not reached. Reached: pass rate on the scored turns the model reached, and their number. $^\dagger$The recorded human drive under the same rule (HUGSIM: nominal start only; AlpaSim: exact binomial interval). $^\ddagger$The lane-keeping model from which our policy is post-trained, with its shipped weights.}
\label{tab:sup-exam-full}
\end{table*}

\paragraph{Turns not reached.}
On Odyssey the six released models do not reach 161 of their scored turns, 16--40\,\% per model against 1\,\% for our policy.
In 114 of them the benchmark has ended the drive for leaving the route, 103 times at a junction (39 at an earlier turn of the list that the model failed), a median 175\,m before the entry; in 15 the time limit of Odyssey ends it, 12 of them garage standing still; and in 32 the car reaches the turn only after the window of the exam, which closes 15\,s after the human driver has left the last listed turn of the route.
No drive is ended by a collision.
Counted on reached turns only, no released model passes half of its Odyssey turns (column Reached).

\paragraph{Home benchmarks.}
WA-JEPA, Alpamayo~1.5 and our policy pass more turns on their home benchmark than on each of the other two, on all turns and on reached turns, with the difference resolved (the turns differ between benchmarks, so the two rates are bootstrapped independently) (for example WA-JEPA on HUGSIM against Odyssey, $+$26.9 points [$+$6.5, $+$45.2] on reached turns).
For garage the home advantage is resolved against neither other benchmark, and for DrivoR, whose authors report on HUGSIM, it is resolved against AlpaSim but not against Odyssey.
Alpamayo~1.5 is trained on the dataset from which the AlpaSim scenes are reconstructed~\cite{alpamayo15_card}; by the partition that this dataset publishes for the Alpamayo models~\cite{pai_av_card}, none of its 57 AlpaSim turns lies in a training clip, 46 lie in held-out clips and 11 cannot be checked.

\paragraph{Selection of the Odyssey turns.}
Of the 113 Odyssey turns, 26 had been studied with our policy before the list was drawn; they meet its rule, which reads no model.
Without them our policy passes 52.0\,\% and is still first, ahead of garage (41.0\,\%), and every released model stays under one half.
Most of its lead comes from turns the other models do not reach: on the 46 turns that every model reached, our policy passes 47.0\,\% and is fifth, behind garage (57.1\,\%), WA-JEPA (54.5\,\%), DrivoR (51.4\,\%) and Alpamayo~1.5 (47.5\,\%), with no paired difference resolved.

\subsection{Straight road}
\label{sup:res-straight}
\begin{figure}[tbp]
\centering
\includegraphics[width=\linewidth]{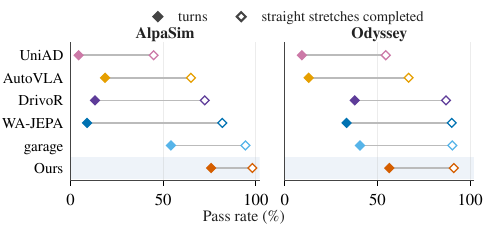}
\caption{\textbf{Straight road against turns.} Turns: pass rate of Table~\ref{tab:exam}. Straight stretches completed: share of stretches of straight road as long as the median turn window (56.5\,m) driven without an off-road event or an at-fault collision. Alpamayo~1.5 has no straight-road read.}
\label{fig:sup-straight}
\end{figure}
Straight road is all driving outside the windows of turns and other manoeuvres, 80--94\,\% of the kilometres of a full AlpaSim or Odyssey read.
Every model completes far more straight stretches than it passes turns (Fig.~\ref{fig:sup-straight}), and per kilometre its failures are 1.4--7.4 times as dense inside the turn windows as on straight road.
For our policy, garage, WA-JEPA and DrivoR the failures are thus specific to turns, while UniAD and AutoVLA fail on a third or more of the straight stretches as well.

\subsection{garage without its route}
\label{sup:garage-noroute}
garage reads its navigation as three target points 40, 60 and 80\,m along the route and has no input for a turn command.
On AlpaSim and Odyssey the benchmark serves a route to every model, and Table~\ref{tab:exam} drives garage as it ships, with the route; HUGSIM serves only a turn command, and there we feed garage the recorded path of the scenario, the most navigation any model receives on that benchmark ($\S$ in Table~\ref{tab:exam}).
garage turns most evenly across the three benchmarks (Sec.~\ref{sec:results}), and a part of its turns could come from being shown where the road goes rather than from driving the junction.
Its authors release the checkpoint of the preceding training stage, trained without target points and published as their unconditional model; driving it with no route on all three benchmarks, and otherwise as the route arm, measures that part.
It is listed as `w/o route' in Table~\ref{tab:sup-exam-full} and Sec.~\ref{sup:why-starts}.

Without the route garage passes 7, 19 and 29 points fewer turns on AlpaSim, Odyssey and HUGSIM (Table~\ref{tab:sup-exam-full}), and the loss is the choice of the turn: the car goes straight on or turns the wrong way at 39\,\% of the HUGSIM starts and 20\,\% of the AlpaSim turns, and on Odyssey it leaves the route or runs out of time before 40\,\% of the junctions.
The benchmark scores fall with the turns: Odyssey RouteDS from 55.2 to 30.7 ($-24.5$ [$-32.1$, $-17.0$] paired), the AlpaSim score from 0.774 to 0.668 ($-0.106$ [$-0.145$, $-0.065$]) and the HUGSIM score from 0.457 to 0.407 ($-0.050$ [$-0.083$, $-0.020$]).
\subsection{How the models fail}
\label{sup:models}

\subsubsection{Turn failures per model}
\label{sup:models-turns}
The most frequent failure of a model is largely the same from one benchmark to the next (Table~\ref{tab:sup-exam-full}): UniAD leaves the road on all three, WA-JEPA and DrivoR do not take the turn on AlpaSim and HUGSIM, Alpamayo~1.5 stalls on both, and our policy leaves the road on HUGSIM and collides at fault on the other two.
On Odyssey, where a failure ends the drive for the turns after it, the turns that are not reached lead for AutoVLA, Alpamayo~1.5 and WA-JEPA.

On the HUGSIM turns the models fail on different sides and in different ways.
Our policy, UniAD, garage, DrivoR and AutoVLA fail on the inside of the turn, with the car more than 0.5\,m inside the recorded path at 78, 81, 71, 67 and 61\,\% of their first departures from the road or collisions: the car cuts the corner.
WA-JEPA and Alpamayo~1.5 fail on the outside (56 and 94\,\%).
Of the failing starts, WA-JEPA does not take the turn in 136 of 200 and DrivoR in 149 of 295, and Alpamayo~1.5 stalls in 202 of 367 and garage in 115 of 310.

\subsubsection{UniAD}
\label{sup:models-uniad}
UniAD plans with a single query: the ego queries of its track and motion heads and one of three learned command vectors are fused into it, it reads the BEV and regresses six points 0.5\,s apart, and a collision optimiser then moves the points away from the predicted occupancy.
The command label of a training sample depends only on whether the last point of the recorded future lies 2\,m to one side.

\begin{figure}[tbp]
  \centering
  \includegraphics[width=\linewidth]{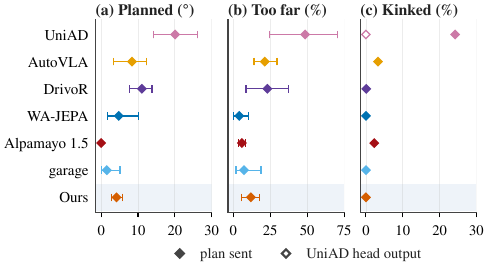}
  \caption{\textbf{Plans of UniAD and the other models on the HUGSIM turns.} Model calls from 10\,m before the junction entry to the exit under the turn command. (a, b)~Calls on which the recorded path turns less than 15\textdegree{} over the length of the plan: heading of the last segment of the 3\,s plan, and share turning more than 10\textdegree{} beyond the path; medians, 95\,\% bootstrap intervals over the 22 junction places. (c)~Plans in which the second difference of consecutive points exceeds 1\,m.}
  \label{fig:sup-uniad-compare}
\end{figure}

\begin{figure*}[tbp]
  \centering
  \includegraphics[width=\linewidth]{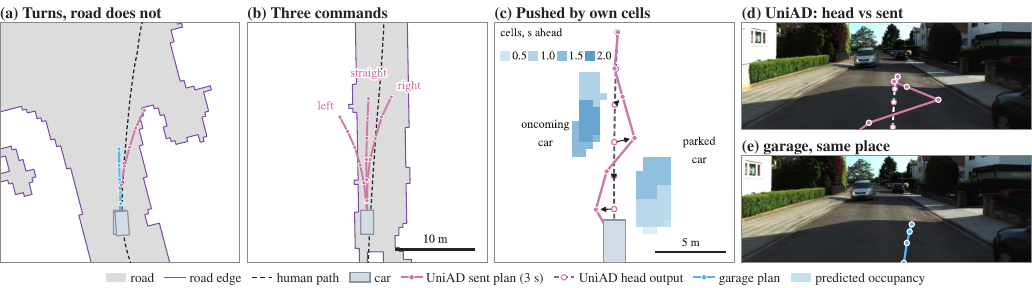}
\caption{\textbf{Plans of UniAD.} One model call per panel; garage at its nearest pose at the same place and start. (b)~One frame on a straight road with the planning head run under each command, in the configuration with the default HUGSIM cameras. (c)~The predicted occupancy each point of the plan in (d) is pushed from (cells within 5\,m of the point, shaded by the future time the point is checked against), with an arrow from each point of the head output to its point in the sent plan. (d, e)~Plans projected onto the ground of the front camera.}
  \label{fig:sup-uniad-plans}
\end{figure*}

Two traits of the plans show on the HUGSIM turns (Figs.~\ref{fig:sup-uniad-compare} and~\ref{fig:sup-uniad-plans}).
Where the road ahead turns little, the plan of UniAD still turns, and turns too far on about half of these steps; the other models plan almost straight there.
On a straight road the three commands turn the plan to three different headings (Fig.~\ref{fig:sup-uniad-plans}b).
The plan it sends is also kinked: the head outputs a smooth curve, and the collision optimiser moves single points of it by more than 1\,m, sideways or along the plan, which no other model does as often (Fig.~\ref{fig:sup-uniad-plans}d, e).
The optimiser moves each point away from the predicted occupancy near it, with no term that couples neighbouring points (Fig.~\ref{fig:sup-uniad-plans}c).
It does not read the drivable area that UniAD itself decodes either: on the calls where it moves a point by more than 1\,m it takes the plan off that area on 4.6\,\% [1.2, 8.2] and brings it back on 0.6\,\% [0.0, 1.6].
The kinks are as frequent in passing turns as in failing ones.
At the entry of sharper corners the plan turns less than the road, as those of WA-JEPA, DrivoR and AutoVLA do.

\subsubsection{AutoVLA and Alpamayo~1.5}
\label{sup:models-vla}
\begin{figure*}[tbp]
  \centering
  \includegraphics[width=\linewidth]{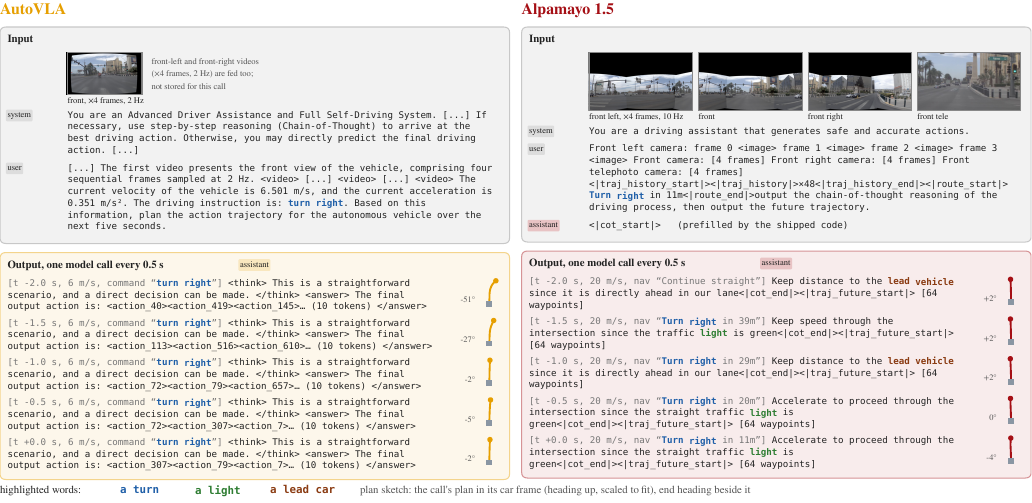}
  \caption{\textbf{What AutoVLA and Alpamayo~1.5 read and write on the approach to one turn on Odyssey.} Input: the newest fed frame of each camera and the prompt as the shipped code builds it, with the values of that call; [\dots] marks cut boilerplate. Output: the text each model generates on five consecutive calls, timed from the junction entry, with a sketch of the plan of the call in its car frame and its end heading.}
  \label{fig:sup-vla-prompt}
\end{figure*}

\begin{figure*}[tbp]
  \centering
  \includegraphics[width=\linewidth]{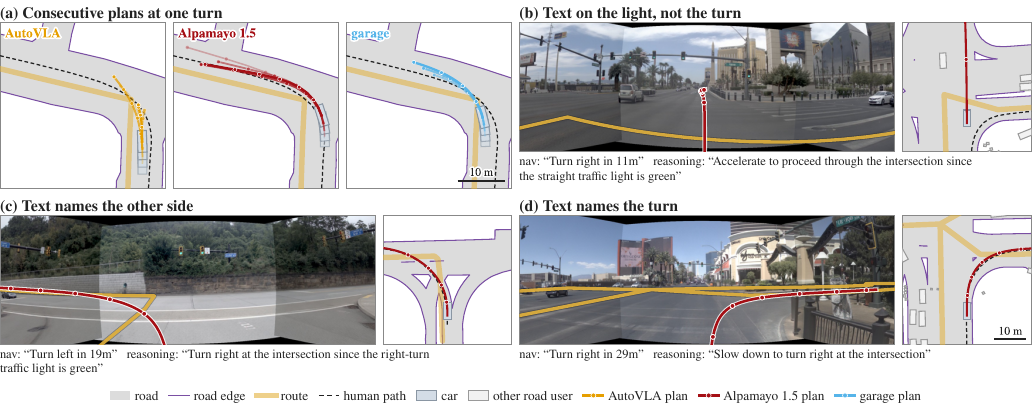}
  \caption{\textbf{Plans and reasoning text of AutoVLA and Alpamayo~1.5 on Odyssey.} (a)~Four consecutive plans of each model at the same turn, from the oldest (faint) to the newest (strong); all three models plan twice a second here. (b--d)~One call of Alpamayo~1.5 each: the front view it is given with its plan and the route projected onto the road, the top view at that moment, and below them the navigation text it is given and the reasoning text it writes, verbatim.}
  \label{fig:sup-vla}
\end{figure*}

AutoVLA is a language model that writes its plan as action tokens and may write reasoning text before them; Alpamayo~1.5 writes a sentence of reasoning before every plan.
The two use the language in different ways (Figs.~\ref{fig:sup-vla-prompt} and~\ref{fig:sup-vla}); Fig.~\ref{fig:sup-vla-prompt} shows what each model reads and writes on the approach to one turn.

As shipped, AutoVLA never writes reasoning: on every one of about 60\,000 calls on the three benchmarks it answers that the scene is straightforward and a direct decision can be made, then writes the action tokens.
Each call is independent of the one before, and each action token is a 0.5\,s motion taken from a fixed codebook, so nothing ties one plan to the next.
Its plans move between calls more than those of the other models except UniAD: in junctions the point 3\,s ahead moves by a median 0.30\,m from one call to the next on Odyssey and 0.66\,m on AlpaSim, against 0.08 and 0.12\,m for garage, which also plans twice a second (Fig.~\ref{fig:sup-vla}a).
The moves are no larger in failing turns than in passing ones, so they are a trait of the structure and do not explain the failures.

Alpamayo~1.5 reasons on every call, but seldom about the route.
On calls within 25\,m of a turn of the route, its text names a turn on 9.5--17\,\% of the calls; the rest speak of the car ahead, the traffic light or a stop (Fig.~\ref{fig:sup-vla}b).
The plan agrees with the text where the text names the turn of the route: the plan then turns that way on 63--91\,\% of the calls, and it goes straight on 60--78\,\% of the calls on which the text names no turn (Fig.~\ref{fig:sup-vla}d).
The text does not steer the plan against the navigation: when it names the other side, the plan mostly turns with the navigation text (Fig.~\ref{fig:sup-vla}c).
Nor does it set the speed: after a text to stop, yield or follow, the plan keeps the current speed as it does after a text to go.
The reasoning thus describes the scene more than it decides the drive, and in the scenes of a turn it mostly describes other things than the turn.

\begin{figure*}[tbp]
  \centering
  \includegraphics[width=\linewidth]{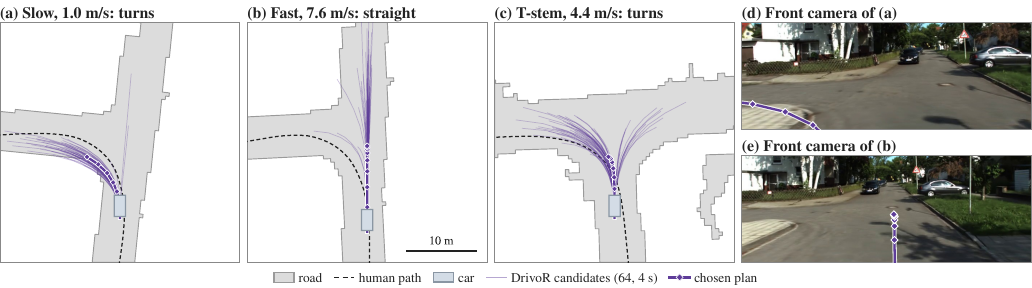}
  \caption{\textbf{Candidates of DrivoR.} The 64 candidate plans and the plan picked by the scorer at one model call per panel. (a, b)~One junction where the road also goes straight, two calls of the same drive. (c)~The stem of a T junction. (d, e)~The front camera of HUGSIM at (a) and (b); DrivoR is fed views at the pose of its training cameras.}
  \label{fig:sup-drivor}
\end{figure*}

\subsubsection{WA-JEPA}
\label{sup:models-wajepa}
\begin{figure}[tbp]
  \centering
  \includegraphics[width=\linewidth]{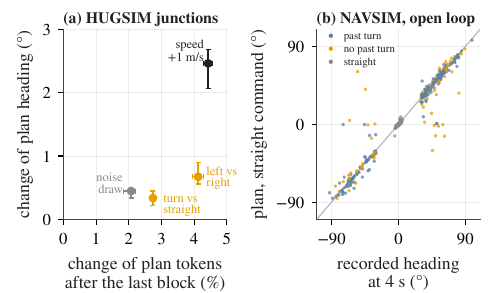}
  \caption{\textbf{Where the turn command of WA-JEPA is lost.} (a)~HUGSIM junction calls under the turn command, re-run with one input changed: relative change of the eight plan tokens after the last block against the change of the heading of the 3\,s plan; medians, 95\,\% bootstrap intervals. (b)~NAVSIM test scenes. Past turn: yaw above 5\textdegree{} over the last 1.5\,s, or slowing.}
  \label{fig:sup-wajepa-cmd}
\end{figure}
WA-JEPA turns where the road turns and drives on where the road goes on, whatever the command.
On HUGSIM and AlpaSim its most frequent failure is a turn it does not take (Table~\ref{tab:sup-exam-full}), and on Odyssey most of the turns it does not reach come after it left the route at an earlier junction.
With the same input, swapping the command from left to right moves the end of its plan by less than one degree at T-junctions, where UniAD and AutoVLA turn their plans by about 80\textdegree{} and DrivoR does so once it has slowed (Sec.~\ref{sup:models-drivor}).

The command reaches the planner and is not read there (Fig.~\ref{fig:sup-wajepa-cmd}a).
It is four of the eight numbers of the ego status; one MLP turns them into a vector that is added once to the eight plan tokens, which then attend together with 4\,096 camera tokens and 8\,192 tokens of the imagined future scene through twelve blocks conditioned on the flow time only.
At the input the command changes the plan tokens as much as a speed change of 1\,m/s.
Both changes shrink through the blocks, and after the last block the difference between left and right is still as large as the speed change and twice the change from another noise draw.
The plan reads the speed and not the command: the speed change turns the 3\,s plan by 2.5\textdegree, the command by less than a degree, no more than another noise draw.

The training data never required the command.
On NAVSIM, where WA-JEPA was trained, 78\,\% of the turns in the training split already show in the past motion of the car, and the command names the side of the turn in 97\,\%.
With the command set to straight the plan still follows the recorded turn on 279 of 300 test turns, those that do not show in the past motion included (Fig.~\ref{fig:sup-wajepa-cmd}b): the images carry the turn, and the command decides 14 of them.
At a junction of a closed-loop drive the images do not carry it where the road also goes on, and the plan follows the road.
On Odyssey, which labels each junction by the exits the road offers, WA-JEPA reaches the routed exit of 86\,\% of the junctions with a single way on and 65\,\% of those with a choice.

The plan also rotates with the car: a heading change seen in the images turns the plan by more than it turned the car, more than twice as much as in our policy, and an outward rotation at the junction entry is the one WA-JEPA fails (Sec.~\ref{sec:mechanism}).

\subsubsection{DrivoR}
\label{sup:models-drivor}
DrivoR regresses 64 candidate plans from its image tokens and its ego state, which holds the speed and a one-hot command, and a learned scorer picks one of them.
Where the road also goes straight it often keeps straight on (Table~\ref{tab:sup-exam-full}).
Its command decides the plan only at low speed (Fig.~\ref{fig:sup-drivor}).
On the NAVSIM samples taken before the driver starts to turn, a straight command removes the turn on most samples logged below 3\,m/s, while from 5\,m/s up most plans still follow the driver with the opposite command; feeding the speed at which DrivoR reaches HUGSIM junctions (7.6\,m/s) removes 17 points of turns, and removing the deceleration changes nothing.
On HUGSIM it reaches the junctions it misses at a median 8.6\,m/s, and there the command moves its plan by 15\textdegree{} or more on 1\,\% of the steps, against a quarter of the steps below 3\,m/s.
Feeding it a speed of 2\,m/s on 110 missed turns at seven junctions makes it turn on 63 and pass 33, against one pass when the fed speed is raised by the same amount.
Slowing the car itself to 2.5\,m/s makes it turn on the same 63 but pass only 13: it stalls inside the turn and swings into the kerb.
The open-loop samples thus come at the speed of a driver who has already slowed for the turn, the one range in which DrivoR reads its command; the cap at the speed of the human driver in Sec.~\ref{sup:why-tested} stays above it.

\subsubsection{Other manoeuvres}
\label{sup:basics}
HUGSIM has no traffic-light state and at most four lane changes per model over its 436 scenarios, and AlpaSim maps traffic lights in 10 of its 433 scenes and holds two U-turns, so neither can score these manoeuvres.
On Odyssey, in one run per model, every model except Alpamayo~1.5, which was not read for this, runs 32--59\,\% of the red lights it meets on its approach (garage 18 of 57 to AutoVLA 32 of 54; our policy 27 of 64) and makes a needed lane change before a junction cleanly in a minority of cases (DrivoR 2 of 32 to garage 7 of 19; our policy 10 of 40).
\section{Why Turns Fail}
\label{sup:why}

\subsection{Explanations tested}
\label{sup:why-tested}
\begin{figure}[tbp]
\centering
\includegraphics[width=\linewidth]{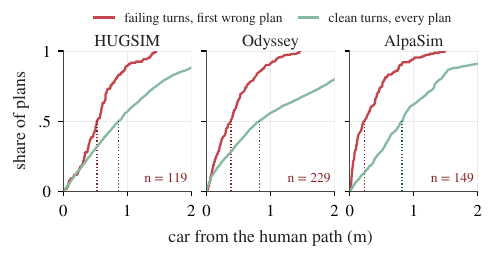}
\caption{\textbf{Where failing turns go wrong.} Cumulative share of the distance of the car from the human path at the first wrong in-junction plan of each failing turn ($n$), and at every in-junction plan of the clean turns; dotted: median.}
\label{fig:why-tested}
\end{figure}

Table~\ref{tab:ruledout} lists the explanations of a failed turn that were tested, each with the change or measure that tests it and its control.
An earlier turn command, target points on the human path and a plan on every simulator tick rescue at most two failing turns more than their placebo.
A plan is wrong inside the junction when its point 2\,s ahead leaves a 1\,m corridor around the human path without return, or goes straight where the path turns.
At the logged poses of 14 turns the plans of garage are rarely wrong, on real and on rendered frames alike, and on 7 of these turns they go wrong more often from the states of its own drive; on every benchmark the first wrong plan of a failing turn comes closer to the human path than the clean turns are driven (Fig.~\ref{fig:why-tested}), so the failure does not wait for states far outside the data.
The car follows its plan: in 23 of 24 model and benchmark cells its speed stays within 0.31\,m/s of the plan over the 20\,m before the entry.
The entry speed is the one explanation left open: four of the five turns of our policy rescued by the speed cap are rescued only to a departure from the route lanes on the road.
Several tests rest on a few dozen turns or fewer; they show that an explanation does not account for the failures of the models tested, and do not exclude a small effect.

\begin{table}[tbp]
\centering
\footnotesize\setlength{\tabcolsep}{3pt}
\begin{tabular}{@{}>{\raggedright\arraybackslash}p{0.31\linewidth} l r r@{}}
\toprule
Explanation & Model, benchmark & Test & Control \\
\midrule
Late turn command~\cite{codevilla2018,navihydra,hua2026command} & DrivoR, HUGSIM & 2/20 & 0/20 \\
 & UniAD, HUGSIM & 0/19 & 0/20 \\
Inaccurate navigation input~\cite{hiddenbiases} & garage, AlpaSim & 6/12 & 4/12 \\
Low replan rate & garage, AlpaSim & 3/12 & 4/12 \\
 & WA-JEPA, AlpaSim & 0/9 & 0/9 \\
Rendering artefacts~\cite{navsim,pseudosim} & garage, AlpaSim & 4/182 & 5/182 \\
Model does not know the road & garage, AlpaSim & 4/182 & 8/69 \\
No recovery data~\cite{dagger,chauffeurnet} & all models & 0.23--0.52\,m & 0.81--0.86\,m \\
\midrule
\emph{Car enters too fast (open)} & Ours, Odyssey & 5/17 & 1/20 \\
 & DrivoR, Odyssey & 1/7 & 1/8 \\
\bottomrule
\end{tabular}
\caption{\textbf{Explanations of a failed turn, tested.} Command, navigation input, replan rate and speed: failed turns rescued by the change (command 40\,m before the entry; target points on the human path; a plan on every simulator tick; entry speed capped at that of the human driver) and by a placebo of matched size (start moved by 5\,cm, or target points handed over 0.2\,s later); rescued: passes the turn (garage: no failure inside the junction; WA-JEPA: no incident). Rendering and road: wrong in-junction plans; recovery data: median distance from the human path, range over benchmarks (Fig.~\ref{fig:why-tested}).}
\label{tab:ruledout}
\end{table}

\subsection{Passing and failing starts}
\label{sup:why-starts}
The read uses the episodes of Table~\ref{tab:exam}, in which no drive is stopped at an off-road plan, and reproduces every HUGSIM cell of it; the arrival state and the separation are read on the 78 turns released 25\,m before the entry, with the lateral offset and heading measured against the recorded path.
Table~\ref{tab:sup-starts} gives every number per model.

\begin{table*}[tbp]
\centering
\footnotesize\setlength{\tabcolsep}{2.4pt}
\emph{(a) Start and arrival}\\[2pt]
\begin{tabular}{@{}l r r l l l l r r r r@{}}
\toprule
 & Mixed & & \multicolumn{2}{c}{Worst 25\,m start} & \multicolumn{2}{c}{40\,m start} & \multicolumn{2}{c}{Lateral offset at entry} & 25 v 40\,m & Heading \\
\cmidrule(lr){4-5}\cmidrule(lr){6-7}\cmidrule(lr){8-9}
Model & turns (\%) & $p$ & start & $\Delta$ pass & $\Delta$ pass & $\Delta$, reached & share left & median (m) & apart (m) & SD (\textdegree) \\
\midrule
UniAD & 34 & 0.62 & lat.\ $+$0.5\,m & $-$0.12{\scriptsize\color{gray}\,[$-$0.28,\,+0.00]} & $-$0.05{\scriptsize\color{gray}\,[$-$0.20,\,+0.11]} & $-$0.03{\scriptsize\color{gray}\,[$-$0.18,\,+0.14]} & 0.48 & 0.28 & 0.17 & 2.5 \\
AutoVLA & 69 & \textbf{0.004} & speed $-$1 & $-$0.16{\scriptsize\color{gray}\,[$-$0.36,\,+0.01]} & +0.11{\scriptsize\color{gray}\,[+0.00,\,+0.27]} & +0.12{\scriptsize\color{gray}\,[+0.00,\,+0.28]} & 0.67 & 0.58 & 0.41 & 6.1 \\
DrivoR & 27 & 0.39 & speed $+$1 & $-$0.01{\scriptsize\color{gray}\,[$-$0.08,\,+0.04]} & \textbf{$-$0.15}{\scriptsize\color{gray}\,[$-$0.32,\,$-$0.02]} & $-$0.07{\scriptsize\color{gray}\,[$-$0.22,\,+0.03]} & 0.37 & 0.21 & 0.31 & 2.6 \\
WA-JEPA & 23 & 0.41 & head.\ $-$3\textdegree & $-$0.06{\scriptsize\color{gray}\,[$-$0.13,\,+0.01]} & \textbf{$-$0.10}{\scriptsize\color{gray}\,[$-$0.23,\,$-$0.01]} & $-$0.06{\scriptsize\color{gray}\,[$-$0.19,\,+0.01]} & 0.51 & 0.28 & 0.30 & 1.1 \\
garage & 37 & 0.07 & lat.\ $-$0.5\,m & $-$0.11{\scriptsize\color{gray}\,[$-$0.29,\,+0.06]} & $-$0.02{\scriptsize\color{gray}\,[$-$0.11,\,+0.08]} & +0.01{\scriptsize\color{gray}\,[$-$0.06,\,+0.11]} & 0.40 & 0.22 & 0.19 & 1.3 \\
\quad w/o route & 15 & 0.89 & lat.\ $-$0.5\,m & $-$0.02{\scriptsize\color{gray}\,[$-$0.10,\,+0.04]} & +0.01{\scriptsize\color{gray}\,[+0.00,\,+0.03]} & +0.03{\scriptsize\color{gray}\,[+0.00,\,+0.07]} & 0.46 & 0.23 & 0.18 & 0.9 \\
Ours & 41 & 0.21 & head.\ $+$3\textdegree & $-$0.07{\scriptsize\color{gray}\,[$-$0.17,\,+0.01]} & $-$0.03{\scriptsize\color{gray}\,[$-$0.07,\,+0.01]} & +0.00{\scriptsize\color{gray}\,[$-$0.04,\,+0.05]} & 0.19 & 0.16 & 0.25 & 3.9--4.4 \\
\bottomrule
\end{tabular}\\[6pt]
\emph{(b) A failing (F) and a passing (P) start of the same turn, and the first failure}\\[2pt]
\begin{tabular}{@{}l r r r r r r r r l r r@{}}
\toprule
 & \multicolumn{2}{c}{Gap at entry (m)} & \multicolumn{2}{c}{Tracks apart (\%)} & \multicolumn{2}{c}{Tracks apart (\%)} & \multicolumn{2}{c}{Plans apart (\%)} & AUC, lateral & \multicolumn{2}{c}{First failure} \\
 & & & \multicolumn{2}{c}{before entry} & \multicolumn{2}{c}{by $+$5\,m} & \multicolumn{2}{c}{before entry} & offset at entry & in junction & past entry \\
\cmidrule(lr){2-3}\cmidrule(lr){4-5}\cmidrule(lr){6-7}\cmidrule(lr){8-9}\cmidrule(lr){11-12}
Model & F--P & P--P & F--P & P--P & F--P & P--P & F--P & P--P & & (\%) & (m) \\
\midrule
UniAD & 0.23 & 0.17 & 2 & 0 & 8 & 1 & 51 & 45 & \textbf{0.80}{\scriptsize\color{gray}\,[0.60,\,0.95]} & 85 & 8.0 \\
AutoVLA & 0.86 & 0.57 & 44 & 36 & 63 & 48 & 94 & 79 & 0.64{\scriptsize\color{gray}\,[0.46,\,0.78]} & 75 & 6.5 \\
DrivoR & 0.37 & 0.18 & 6 & 1 & 51 & 27 & 65 & 35 & \textbf{0.31}{\scriptsize\color{gray}\,[0.14,\,0.43]} & 75 & 7.5 \\
WA-JEPA & 0.40 & 0.21 & 4 & 0 & 45 & 4 & 48 & 12 & 0.74{\scriptsize\color{gray}\,[0.48,\,0.98]} & 71 & 10.0 \\
garage & 0.23 & 0.10 & 23 & 0 & 29 & 3 & 29 & 7 & 0.67{\scriptsize\color{gray}\,[0.49,\,1.00]} & 90 & 10.0 \\
\quad w/o route & 0.33 & 0.22 & 37 & 19 & 54 & 24 & 40 & 31 & 0.59{\scriptsize\color{gray}\,[0.07,\,1.00]} & 86 & 9.0 \\
Ours & 0.28 & 0.21 & 11 & 5 & 48 & 31 & 59 & 34 & \textbf{0.74}{\scriptsize\color{gray}\,[0.65,\,0.85]} & 91 & 8.0 \\
Alpamayo~1.5 & -- & -- & -- & -- & -- & -- & -- & -- & -- & 94 & 13.0 \\
\bottomrule
\end{tabular}
\caption{\textbf{The eight HUGSIM starts of a turn, per model.} Mixed turns: turns with both passing and failing starts. $p$: permutation test over the seven starts released 25\,m before the entry, relabelling the starts once per scene. $\Delta$ pass: paired difference in pass rate against the nominal start with its 95\,\% scene-bootstrap interval; `reached', without the runs that end on the 15\,m added by the 40\,m start. Share left: of the 0.5\,m lateral offset of a start; median: absolute lateral offset at the entry. 25 v 40\,m apart: distance at the entry between the two unperturbed starts. Heading SD: across the starts of a turn at the entry (Ours: range over three seeds). Apart: 1\,m or more between the tracks, or between the plan points 2\,s ahead. AUC: absolute lateral offset at the entry, failing against passing starts. First failure: of the failing starts that leave the road or collide, share whose first such event lies inside the junction or past its exit, and its median distance past the entry. Bold: interval excludes zero (AUC: 0.5), $p<0.05$. Alpamayo~1.5 has no start-variant read.}
\label{tab:sup-starts}
\end{table*}

\paragraph{Pass rate by start.}
Pooled over the seven rows, each perturbed 25\,m start lies within $-$0.028 to $+$0.010 of the nominal start, and every interval includes zero; per model no 25\,m start is resolved below the nominal one, and the permutation test finds an effect of the start for AutoVLA alone.
The 40\,m start, which adds 15\,m of approach driven by the model, costs DrivoR and WA-JEPA a resolved share of their turns, and the loss is no longer resolved once the runs that end on the added approach are left out.

\paragraph{State on arrival.}
Heading and speed offsets are largely gone 10\,m before the entry for every model.
Of the lateral offset, our policy keeps a fifth at the entry, AutoVLA two thirds and the other models a third to a half.
The heading of our policy at the entry varies more across the starts of a turn than that of any other model except AutoVLA.

\paragraph{Failing against passing starts of one turn.}
On the turns where some starts pass and others fail, a failing and a passing start arrive at the entry further apart than two passing starts for every model, yet their tracks are rarely 1\,m apart before the entry, garage with and without its route (23 and 37\,\%) and AutoVLA excepted.
They part inside the junction: 5\,m past the entry the tracks of a failing and a passing start are 1\,m apart in 29--63\,\% of the pairs for every model except UniAD (8\,\%), more often than those of two passing starts, and most clearly so for WA-JEPA and garage (45 and 29\,\% against 4 and 3\,\%).
The plan points 2\,s ahead part earlier, since at the entry that point lies 8--12\,m into the junction.
The first failure lies at the junction as well: for every model 71--94\,\% of the failing starts that leave the road or collide do so first inside the junction or past its exit.
The passing starts are off the human path themselves, by a median 0.27--0.65\,m and 2--6\textdegree{} per model at the entry, so an offset from the human path does not by itself fail a turn.
The lateral offset at the entry separates failing from passing starts for our policy and UniAD; for DrivoR, on 10 turns, the failing starts lie closer to the path, and for the other models the interval includes 0.5.
The state on arrival thus decides the turn in part for our policy and UniAD and gives no evidence for the other models.
AutoVLA differs along the whole approach: it keeps the most of its offset, and its starts are 1\,m apart before the entry in 36--44\,\% of the pairs whatever the outcome.

\subsection{Interventions}
\label{sup:why-interventions}
Three interventions were registered before any episode: a rotation of the car at the junction entry, a delay of the first fresh plan after such a rotation, and an earlier command or a capped entry speed on failed turns (Table~\ref{tab:ruledout}).
Each runs the model unchanged, on turns it otherwise passes for the first two and on turns it fails for the third, in a copy of the simulator whose empty intervention repeats the stored episode value for value, and each has a placebo of matched size.
All are scored by the pass rule of the paper; Table~\ref{tab:sup-plan} gives every count of the interventions on the car and on the plan.

\begin{table}[tbp]
\centering
\footnotesize\setlength{\tabcolsep}{1.4pt}
\begin{tabular}{@{}lcc@{\hspace{3pt}}ccc@{}}
\toprule
 & \multicolumn{2}{c}{Change of the car} & garage & Ours & WA-JEPA \\
\cmidrule(r{6pt}){2-3}
\emph{(a) Passed turns: failed} & heading & position & {\scriptsize AlpaSim} & {\scriptsize AlpaSim} & {\scriptsize HUGSIM} \\
\midrule
Plan held, 1.8\,m in & 3--6\textdegree & $\le$0.1\,m & 6\,/\,16 & 11\,/\,27 & 0\,/\,44 \\
\quad 0.6\,m in & 0.7--2\textdegree & $<$0.05\,m & 1\,/\,17 & 2\,/\,27 & 0\,/\,44 \\
\quad 1.8\,m out & 3--5\textdegree & $\le$0.1\,m & 1\,/\,17 & 3\,/\,27 & 9\,/\,44 \\
\addlinespace
Shift & 0 & 5\,cm & 0\,/\,34 & 1\,/\,54 & 0\,/\,88 \\
Rotation, in & 2\textdegree & 0 & 1\,/\,17 & 1\,/\,27 & 0\,/\,44 \\
 & 4\textdegree & 0 & 3\,/\,17 & 5\,/\,27 & 0\,/\,44 \\
 & 6\textdegree & 0 & 4\,/\,17 & 6\,/\,27 & 0\,/\,44 \\
Rotation, out & 4\textdegree & 0 & 0\,/\,17 & 0\,/\,26 & 14\,/\,44 \\
\midrule
 & \multicolumn{2}{c}{Plan corrected in} & garage & Ours & Odyssey \\
\cmidrule(r{6pt}){2-3}
\emph{(b) Failed turns: clean} & heading & offset & {\scriptsize AlpaSim} & {\scriptsize AlpaSim} & {\scriptsize 3 models} \\
\midrule
Heading error & yes & no & 7\,/\,10 & 9\,/\,19 & 20\,/\,42 {\scriptsize (5,\,9,\,6)} \\
Heading, offset & yes & yes & 6\,/\,10 & 9\,/\,19 & 17\,/\,42 {\scriptsize (4,\,8,\,5)} \\
Control & random & no & 0\,/\,10 & 2\,/\,19 & 1\,/\,42 {\scriptsize (0,\,1,\,0)} \\
\bottomrule
\end{tabular}
\caption{\textbf{Every intervention at the junction.} Change of the car: for a plan held 0.5\,s, the range of the medians per model when control returns to the model. Plan held: 1.8 or 0.6\,m toward the inside or the outside of the turn (Sec.~\ref{sup:why-interventions}). Shift and rotation: at the junction entry; the shift is the placebo of the rotation. Odyssey: failures by running wide, 14 per model; in parentheses garage, WA-JEPA and Ours. Ours: one seed.}
\label{tab:sup-plan}
\end{table}

\paragraph{A heading error at the entry.}
A heading error of 2\textdegree{} is mostly survived, and the failing side depends on the model: garage and our policy fail toward the inside, WA-JEPA toward the outside, mostly by turning the wrong way, by a static collision or off the road on the outside.
By the line registered before the runs the inward rotation supports a causal role of the heading error for our policy on AlpaSim and does not support it for WA-JEPA on HUGSIM.

\paragraph{Correcting the plan.}
The plan corrections of Sec.~\ref{sec:why} run in the same patched copy of AlpaSim, which with an empty intervention repeats the benchmark run value for value on 17 of 17 turns of garage and 27 of 27 of our policy; on Odyssey they run in the simulator process, and 9 of 135 runs that differed from their source before the turn are left out.

\paragraph{A wrong plan held for half a second.}
A fourth intervention, run outside the registered three in the same harness, assumes no kind of error: inside the junction the plan is replaced for 0.5\,s by one whose point 2\,s ahead lies 1.8\,m toward the inside of the turn, the size of the wrong plans of garage in a junction, and control then returns to the model.
The failures of garage all come after control returns, and the nine of WA-JEPA under the outward plan lie at three junctions.
For garage and our policy none of the first plans after the return steers back.
The error such a plan leaves is thus mainly one of heading, which the registered rotation applies alone, and it fails turns mostly on the same side as the rotation for each model.

\begin{figure}[tbp]
\centering
\includegraphics[width=\linewidth]{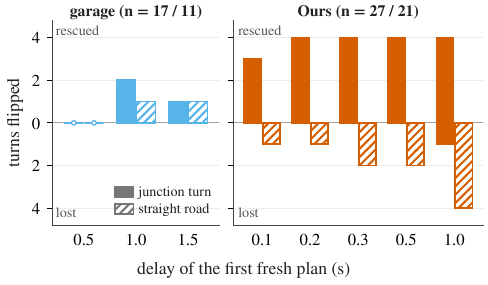}
\caption{\textbf{Fresh plans withheld after a rotation} (AlpaSim). Turns whose outcome changes against no delay. Straight road: the same delay at a straight point of the same scenes. $n$: junction / straight-road turns; Ours: one seed.}
\label{fig:sup-delay}
\end{figure}
\paragraph{Delaying the first fresh plan.}
After a rotation of 4\textdegree{} toward the inside at the junction entry on AlpaSim, the plans the model issues during the next $k$ calls are withheld and the controller follows the clean path of the model from before the rotation, turned with the car (Fig.~\ref{fig:sup-delay}).
Without delay garage passes 14 of 17 rotated turns and our policy 22 of 27; withholding the fresh plans fails almost no turn and rescues some, while on straight road it costs our policy up to 4 of 21 turns at 1.0\,s and garage none.
The registered reading is `not supporting' for each model and pooled: the association between an early first correction and a pass is not caused by the timing of the fresh plans after a deviation.
\section{Training for Turns: Details and All Fine-tunes}
\label{sup:remedy}

A paired difference is a difference in pass rate on the turns scored for both checkpoints, with its 95\,\% scene-bootstrap interval (Sec.~\ref{sup:bench-stat} (\protect\nameref{sup:bench-stat})).

\subsection{Method}
\label{sup:remedy-chain}
\begin{figure*}[tbp]
\centering
\includegraphics[width=\linewidth]{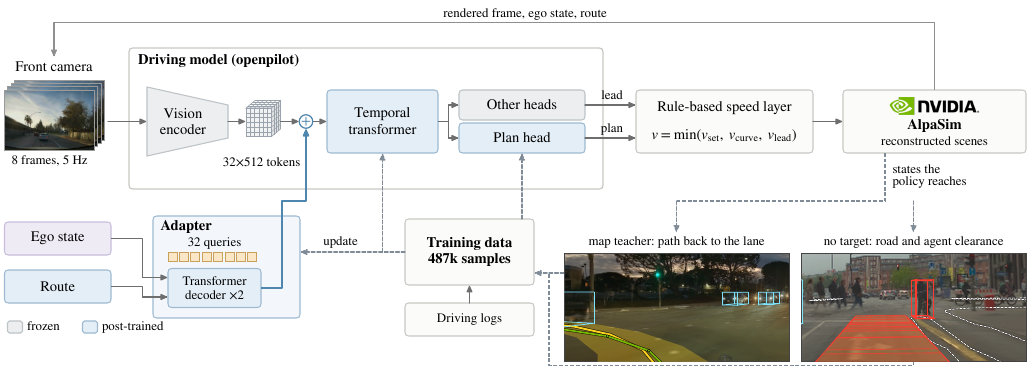}
\caption{\textbf{Our policy and its closed-loop training.} The vision encoder emits 32 tokens of width 512 for each pair of consecutive frames; the adapter adds to them a term computed from the ego state and the route by a transformer decoder with 32 learned queries. $v_\mathrm{set}$, $v_\mathrm{curve}$, $v_\mathrm{lead}$: set speed, limit from the curvature of the plan, limit from the lead vehicle. Dashed: post-training. Left image: a state the policy reaches in AlpaSim, with its plan (yellow), the path back to the lane from a privileged map teacher (green) and other road users (cyan). Right image: a state with no target, where planned positions inside the clearance margin (red) are penalised; red box: a rider; white dashes: road edges.}
\label{fig:sup-policy}
\end{figure*}

\begin{table*}[tbp]
\centering
\footnotesize\resizebox{\textwidth}{!}{\setlength{\tabcolsep}{3pt}
\begin{tabular}{@{}ll@{\hspace{6pt}}r@{\hspace{3pt}}l@{\hspace{4pt}}r@{\hspace{3pt}}l@{\hspace{4pt}}r@{\hspace{4pt}}r@{\hspace{6pt}}r@{\hspace{3pt}}l@{\hspace{4pt}}r@{\hspace{3pt}}l@{\hspace{4pt}}r@{\hspace{4pt}}r@{}}
\toprule
 &  & \multicolumn{6}{c}{Odyssey (113)} & \multicolumn{6}{c}{HUGSIM (99 $\times$ 8 starts)} \\
\cmidrule(lr){3-8}\cmidrule(l){9-14}
Arm & Training & \multicolumn{2}{c}{Pass $\uparrow$} & \multicolumn{2}{c}{$\Delta$ to the control} & \multicolumn{1}{c}{6k} & \multicolumn{1}{c}{earlier} & \multicolumn{2}{c}{Pass $\uparrow$} & \multicolumn{2}{c}{$\Delta$ to the control} & \multicolumn{1}{c}{6k} & \multicolumn{1}{c}{earlier} \\
\midrule
Control (recipe continued) & fine-tune & 60.2 & {\scriptsize\color{gray}[50,\,70]} &  &  &  &  & 52.0 & {\scriptsize\color{gray}[34,\,67]} &  &  &  &  \\
Control, 6,000 steps & fine-tune & 54.0 & {\scriptsize\color{gray}[44,\,64]} & \textbf{$-$0.092} & {\scriptsize\color{gray}[$-$0.164,\,$-$0.020]} &  &  & 51.1 & {\scriptsize\color{gray}[35,\,66]} & $+$0.0002 & {\scriptsize\color{gray}[$-$0.057,\,$+$0.069]} &  &  \\
Earlier control & fine-tune & 52.5 & {\scriptsize\color{gray}[42,\,62]} & \textbf{$-$0.090} & {\scriptsize\color{gray}[$-$0.162,\,$-$0.020]} & 0.000 &  & 49.2 & {\scriptsize\color{gray}[33,\,64]} & $-$0.027 & {\scriptsize\color{gray}[$-$0.097,\,$+$0.046]} & $-$0.026 &  \\
Seed 1 (parent of all arms) & -- & 57.3 & {\scriptsize\color{gray}[47,\,66]} & $-$0.050 & {\scriptsize\color{gray}[$-$0.125,\,$+$0.022]} & $+$0.051 & $+$0.050 & 51.4 & {\scriptsize\color{gray}[35,\,66]} & $-$0.020 & {\scriptsize\color{gray}[$-$0.094,\,$+$0.050]} & $-$0.020 & $+$0.007 \\
\midrule
\multicolumn{14}{@{}l}{\emph{Recovery data, accurate route in the rollouts}} \\
\quad 3 rounds of kicks & fine-tune & 57.7 & {\scriptsize\color{gray}[47,\,68]} & $-$0.063 & {\scriptsize\color{gray}[$-$0.161,\,$+$0.037]} & 0.000 & $+$0.032 & 34.2 & {\scriptsize\color{gray}[18,\,50]} & \textbf{$-$0.168} & {\scriptsize\color{gray}[$-$0.273,\,$-$0.063]} & \textbf{$-$0.167} & \textbf{$-$0.138} \\
\quad 3 rounds of kicks & from the start & 54.0 & {\scriptsize\color{gray}[43,\,65]} & $-$0.093 & {\scriptsize\color{gray}[$-$0.190,\,0.000]} & $-$0.032 & $-$0.021 & 39.4 & {\scriptsize\color{gray}[24,\,55]} & \textbf{$-$0.129} & {\scriptsize\color{gray}[$-$0.239,\,$-$0.019]} & \textbf{$-$0.128} & \textbf{$-$0.099} \\
\quad 3 rounds, route shift (2/3) & from the start & --$^{\ast}$ &  & -- &  &  &  & 33.8 & {\scriptsize\color{gray}[18,\,51]} & \textbf{$-$0.174} & {\scriptsize\color{gray}[$-$0.310,\,$-$0.038]} & \textbf{$-$0.179} & \textbf{$-$0.146} \\
\midrule
\multicolumn{14}{@{}l}{\emph{Route-perturbed recovery data (on top of the 3 rounds above)}} \\
\quad 6 rounds & fine-tune & 62.2 & {\scriptsize\color{gray}[51,\,73]} & $-$0.010 & {\scriptsize\color{gray}[$-$0.104,\,$+$0.086]} & $+$0.065 & $+$0.074 & 43.6 & {\scriptsize\color{gray}[27,\,61]} & $-$0.088 & {\scriptsize\color{gray}[$-$0.221,\,$+$0.049]} & $-$0.086 & $-$0.060 \\
\quad 12 rounds & fine-tune & 66.3 & {\scriptsize\color{gray}[56,\,76]} & $+$0.021 & {\scriptsize\color{gray}[$-$0.067,\,$+$0.105]} & \textbf{$+$0.106} & \textbf{$+$0.117} & 47.1 & {\scriptsize\color{gray}[32,\,63]} & $-$0.049 & {\scriptsize\color{gray}[$-$0.163,\,$+$0.077]} & $-$0.048 & $-$0.025 \\
\quad 6 rounds, half of each batch & fine-tune & 61.8 & {\scriptsize\color{gray}[53,\,70]} & $+$0.010 & {\scriptsize\color{gray}[$-$0.081,\,$+$0.101]} & $+$0.083 & \textbf{$+$0.103} & 34.7 & {\scriptsize\color{gray}[17,\,54]} & $-$0.166 & {\scriptsize\color{gray}[$-$0.352,\,$+$0.022]} & \textbf{$-$0.169} & $-$0.146 \\
\quad 12 rounds, route shift (2/3) & fine-tune & --$^{\ast}$ &  & -- &  &  &  & 47.1 & {\scriptsize\color{gray}[31,\,64]} & $-$0.052 & {\scriptsize\color{gray}[$-$0.186,\,$+$0.079]} & $-$0.051 & $-$0.028 \\
\quad 12 rounds, road-level defects & fine-tune & 59.6 & {\scriptsize\color{gray}[49,\,69]} & $-$0.030 & {\scriptsize\color{gray}[$-$0.129,\,$+$0.067]} & $+$0.051 & $+$0.051 & 55.6 & {\scriptsize\color{gray}[39,\,71]} & $+$0.036 & {\scriptsize\color{gray}[$-$0.070,\,$+$0.152]} & $+$0.039 & $+$0.066 \\
\quad 12 rounds & from the start & 67.6 & {\scriptsize\color{gray}[58,\,76]} & $+$0.061 & {\scriptsize\color{gray}[$-$0.029,\,$+$0.156]} & \textbf{$+$0.126} & \textbf{$+$0.146} & 45.5 & {\scriptsize\color{gray}[29,\,62]} & $-$0.067 & {\scriptsize\color{gray}[$-$0.209,\,$+$0.072]} & $-$0.076 & $-$0.053 \\
\quad 12 rounds, route shift (2/3) & from the start & 66.0 & {\scriptsize\color{gray}[55,\,76]} & $+$0.051 & {\scriptsize\color{gray}[$-$0.039,\,$+$0.143]} & \textbf{$+$0.147} & \textbf{$+$0.146} & 50.7 & {\scriptsize\color{gray}[34,\,67]} & $-$0.030 & {\scriptsize\color{gray}[$-$0.126,\,$+$0.074]} & $-$0.027 & $+$0.005 \\
\quad 12 rounds, route shift (all) & from the start & --$^{\ast}$ &  & -- &  &  &  & 46.0 & {\scriptsize\color{gray}[29,\,63]} & $-$0.057 & {\scriptsize\color{gray}[$-$0.180,\,$+$0.076]} & $-$0.059 & $-$0.030 \\
\quad 12 + 6 rounds, approach kicks$^{\ddagger}$ & fine-tune & 63.4 & {\scriptsize\color{gray}[53,\,73]} & 0.000 & {\scriptsize\color{gray}[$-$0.083,\,$+$0.086]} & $+$0.093 & \textbf{$+$0.092} & 52.0 &  & $-$0.003 & {\scriptsize\color{gray}[$-$0.116,\,$+$0.120]} & $-$0.004 & $+$0.024 \\
\bottomrule
\end{tabular}}
\caption{\textbf{Every one-seed fine-tune of our policy on LaWanle} (one training run each, all from seed 1). Pass rate (\%) with its 95\,\% scene-bootstrap interval and the paired difference in pass rate to the control (a fraction of the turns), which continues the training from seed 1 with no new data for the 12{,}000 steps of the arms; bold where the interval excludes zero. Rounds are collection rounds (with a perturbed route in the lower block); an arm draws a quarter of each batch from the added samples unless stated. Route shift: the fed route is also shifted on two thirds (2/3) or all of the other closed-loop samples of the recipe. $^{\ast}$No Odyssey read under the pass rule of the paper; under the earlier rule in which the first event decides, their differences to the earlier control are $-0.009$ [$-0.095$, $+0.075$], $+0.054$ [$-0.029$, $+0.136$] and $+0.098$ [$0.000$, $+0.196$], in the order of the table. HUGSIM turns are judged on the car. 6k, earlier: the difference to the control of 6{,}000 steps and to the earlier control, bold where its interval excludes zero. $^{\ddagger}$Twelve rounds and six with kicks on the approach: seed 1 of the three-seed fine-tune; its HUGSIM interval was not computed.}
\label{tab:sup-remedy-arms}
\end{table*}

\begin{table*}[tbp]
\centering
\footnotesize\setlength{\tabcolsep}{3pt}
\begin{tabular}[t]{@{}l l l l l@{}}
\toprule
 & & Odyssey (113) & HUGSIM (99 $\times$ 8) & AlpaSim (57) \\
Fine-tune & Against & $\Delta$ paired & $\Delta$ paired & pass, \% \\
\midrule
\multicolumn{5}{@{}l}{\emph{Twelve rounds and six with kicks on the approach}} \\
\quad seed 0 & parent & \textbf{$+$0.131} {\scriptsize\color{gray}[$+$0.040,\,$+$0.227]} & $+$0.013 {\scriptsize\color{gray}[$-$0.122,\,$+$0.151]} & 75.0 v 75.0 \\
\quad seed 1 & parent & $+$0.040 {\scriptsize\color{gray}[$-$0.048,\,$+$0.130]} & $+$0.016 {\scriptsize\color{gray}[$-$0.083,\,$+$0.137]} & 72.7 v 72.2 \\
\quad seed 2 & parent & \textbf{$+$0.124} {\scriptsize\color{gray}[$+$0.020,\,$+$0.227]} & $-$0.066 {\scriptsize\color{gray}[$-$0.200,\,$+$0.068]} & 75.5 v 80.4 \\
\quad three seeds & parent & \textbf{$+$0.082} {\scriptsize\color{gray}[$+$0.007,\,$+$0.157]} & $-$0.006 {\scriptsize\color{gray}[$-$0.103,\,$+$0.111]} & 74.4 v 75.9 \\
\quad three seeds & 6k controls & \textbf{$+$0.090} {\scriptsize\color{gray}[$+$0.018,\,$+$0.162]} & $-$0.048 {\scriptsize\color{gray}[$-$0.145,\,$+$0.063]} & -- \\
\quad seed 1 & 12k control & $+$0.000 {\scriptsize\color{gray}[$-$0.083,\,$+$0.086]} & $-$0.003 {\scriptsize\color{gray}[$-$0.116,\,$+$0.120]} & -- \\
\addlinespace[3pt]
\multicolumn{5}{@{}l}{\emph{Four rounds}} \\
\quad seed 0 & parent & $+$0.072 {\scriptsize\color{gray}[$-$0.034,\,$+$0.183]} & $-$0.079 {\scriptsize\color{gray}[$-$0.240,\,$+$0.073]} & 67.3 v 75.0 \\
\quad seed 1 & parent & $+$0.092 {\scriptsize\color{gray}[$-$0.011,\,$+$0.202]} & $+$0.016 {\scriptsize\color{gray}[$-$0.095,\,$+$0.140]} & 69.8 v 72.2 \\
\quad seed 2 & parent & \textbf{$+$0.122} {\scriptsize\color{gray}[$+$0.035,\,$+$0.211]} & $-$0.034 {\scriptsize\color{gray}[$-$0.134,\,$+$0.084]} & 71.7 v 80.4 \\
\quad three seeds & parent & \textbf{$+$0.094} {\scriptsize\color{gray}[$+$0.014,\,$+$0.179]} & $-$0.031 {\scriptsize\color{gray}[$-$0.117,\,$+$0.073]} & 69.6 v 75.9 {\scriptsize($-$0.065)} \\
\quad three seeds & 6k controls & \textbf{$+$0.105} {\scriptsize\color{gray}[$+$0.031,\,$+$0.179]} & $-$0.074 {\scriptsize\color{gray}[$-$0.163,\,$+$0.026]} & -- \\
\quad seed 1 & 12k control & $+$0.051 {\scriptsize\color{gray}[$-$0.044,\,$+$0.146]} & $-$0.008 {\scriptsize\color{gray}[$-$0.097,\,$+$0.089]} & $-$0.078 {\scriptsize\color{gray}[$-$0.196,\,$+$0.020]} \\
\bottomrule
\end{tabular}
\caption{\textbf{The two fine-tunes run from three seeds, on the turns.} Paired difference in pass rate to the seed a run starts from (parent), to the controls of 6{,}000 steps of the same seeds, and, for seed 1, to the control of 12{,}000 steps; AlpaSim: pass rate of the fine-tune against its parent, not paired (four rounds over three seeds: the paired difference in parentheses); --: not read. Three seeds: Odyssey and AlpaSim, the mean of the seed pairs per turn; HUGSIM, the seed means paired start by start.}
\label{tab:sup-remedy-seeds}
\end{table*}

\begin{table}[tbp]
\centering
\footnotesize\setlength{\tabcolsep}{2.5pt}
\begin{tabular}{@{}l r r r r@{}}
\toprule
 & & \multicolumn{3}{c}{Rows per batch of 128} \\
\cmidrule(l){3-5}
 & & Offline & \multicolumn{2}{c}{Closed loop, steps} \\
\cmidrule(l){4-5}
Samples & Distinct & 0--10k & 0--10k & 10k--19.5k \\
\midrule
Human drive & 103{,}288 & 86.25 & 77.25 & 53.25 \\
\quad retention & (same) & 28.75 & 25.75 & 17.75 \\
Perturbed pose & 69{,}572 & 13 & 13 & 13 \\
Self-correction & 273{,}423 & 0 & 12 & 12 \\
Target-free & 40{,}944 & 0 & 0 & 32 \\
\bottomrule
\end{tabular}
\caption{\textbf{Training samples of our policy.} Rows per batch: expected number of the 128 rows of a batch drawn from each kind of sample; closed-loop samples exist only after the offline stage, and the target-free ones enter at step 10k. Human drive and perturbed pose: nuPlan logs labelled with the path of the driver, the perturbed pose seen from a pose shifted off the recorded path. Retention: the human-drive frames again, with the route and ego inputs zeroed, trained toward the plan of the shipped model. Self-correction: states reached in AlpaSim, labelled by the map teacher. Target-free: states reached in nuPlan and AlpaSim with no path label, trained only to keep the plan on the road and clear of other agents.}
\label{tab:sup-data}
\end{table}

Our policy is the large driving model of openpilot~\cite{openpilot}, Cinque Terre v3 with 382\,M parameters (384.7\,M with the adapter), with its vision encoder frozen, a small adapter that reads the route and the ego state, and the temporal transformer and plan head behind them post-trained (Fig.~\ref{fig:sup-policy}); the shipped openpilot of every table is this model.
Every training starts from the shipped weights.
The offline stage trains for 10{,}000 steps on human drives from nuPlan~\cite{nuplan} and on perturbed poses of the same logs (Table~\ref{tab:sup-data}).
Closed-loop data come from rounds of collection, as in DAgger~\cite{dagger}: the policy of the latest training drives the 719 AlpaSim training scenes, which share no scene with any evaluation, a map teacher labels each state it reaches with a lateral path back to the lane (self-correction), and the next training adds these samples; one round runs in nuPlan scenes and four in AlpaSim.
The final training of 19{,}500 steps uses every round and is run from three seeds. Of the 384.7\,M parameters 19.3\,M are trained, with a learning rate of $3\times10^{-5}$ ($3\times10^{-4}$ for the adapter), a cosine schedule with 300 warm-up steps that restarts at a tenth of the rate when the target-free samples enter at step 10{,}000, weight decay 0.01 and gradients clipped at 1.0; the route input is dropped on a quarter of the rows.
Collecting the closed-loop data takes about 28 GPU-hours on RTX~6000D cards and the trainings about 8, of which the final training of one seed takes about 1.1; a change to the training that reuses the collected data costs the training only.

\paragraph{Recovery data and route perturbation.}
\label{sup:remedy-data}
\emph{Recovery data}~\cite{chauffeurnet,codevilla2019} come from three rounds of closed-loop rollouts in AlpaSim training scenes in which the car is kicked in heading or lateral position on the approach to a junction and inside it (inside the junction up to 2\textdegree{} and 0.875\,m); the 9{,}091 states within 1.5\,s after a kick are labelled by the teacher, and a fine-tune draws 32 of the 128 samples of each batch from them.
\emph{Route-perturbed recovery data} are further rounds on 129 training scenes in which the route fed to the policy is accurate in a third of the episodes and otherwise shifted as a whole by up to 3\,m across and along the road and jittered (median error at the car 1.75\,m), while the teacher labels the way back to the true lane; kicks are larger (2--15\textdegree{} or 0.5--2\,m) and confined to the junction.

\paragraph{Overlap with Odyssey.}
The nuPlan logs of the training draw on the same logs as Odyssey: 68 of the 100 Odyssey scenes contain frames of the training samples.
Against UniAD, AutoVLA, DrivoR, WA-JEPA and garage, which are trained on other logs, our policy shows no advantage on these scenes (difference in differences $+$1.41 RouteDS [$-$11.46, $+$15.18]), a test that rules out only an advantage above about 15 points; on the 32 other scenes it scores 47.68 against 51.30 on all 100.
On the turns the same comparison gives $+$0.03 [$-$0.09, $+$0.16] in the share of turns without incident.

\begin{table}[tbp]
\centering
\footnotesize\setlength{\tabcolsep}{3pt}
\begin{tabular}[t]{@{}l@{\hspace{4pt}}c@{\hspace{4pt}}r@{\hspace{4pt}}r@{\hspace{4pt}}r@{\hspace{3pt}}l@{}}
\toprule
Full benchmark & Seed & parent & fine-tune & \multicolumn{2}{c}{$\Delta$ paired} \\
\midrule
AlpaSim score & 0 & 0.908 & 0.909 & $+$0.002 & {\scriptsize\color{gray}[$-$0.017,\,$+$0.021]} \\
 & 1 & 0.912 & 0.907 & $-$0.005 & {\scriptsize\color{gray}[$-$0.027,\,$+$0.016]} \\
 & 2 & 0.902 & 0.898 & $-$0.004 & {\scriptsize\color{gray}[$-$0.030,\,$+$0.021]} \\
 & mean & 0.907 & 0.905 & $-$0.003 & {\scriptsize\color{gray}[$-$0.020,\,$+$0.014]} \\
\addlinespace[3pt]
Odyssey RouteDS & 0 & 48.74 & 51.98 & $+$3.25 & {\scriptsize\color{gray}[$-$2.45,\,$+$9.17]} \\
 & 1 & 51.30 & 51.01 & $-$0.29 & {\scriptsize\color{gray}[$-$5.50,\,$+$4.76]} \\
 & 2 & 48.09 & 52.90 & $+$4.82 & {\scriptsize\color{gray}[$-$0.76,\,$+$10.48]} \\
 & mean & 49.37 & 51.96 & $+$2.59 & {\scriptsize\color{gray}[$-$1.43,\,$+$6.67]} \\
\addlinespace[3pt]
HUGSIM HD-Score & 0 & 0.455 & 0.466 & $+$0.010 & {\scriptsize\color{gray}[$-$0.014,\,$+$0.036]} \\
 & 1 & 0.462 & 0.476 & $+$0.015 & {\scriptsize\color{gray}[$-$0.011,\,$+$0.041]} \\
 & 2 & 0.453 & 0.473 & $+$0.022 & {\scriptsize\color{gray}[$-$0.002,\,$+$0.048]} \\
 & mean & 0.456 & 0.470 & $+$0.016 & {\scriptsize\color{gray}[$-$0.002,\,$+$0.036]} \\
\addlinespace[3pt]
NAVSIM PDMS & 0 & 88.26 & 88.25 & $-$0.02 & {\scriptsize\color{gray}[$-$0.27,\,$+$0.24]} \\
 & 1 & 88.61 & 88.13 & \textbf{$-$0.48} & {\scriptsize\color{gray}[$-$0.80,\,$-$0.17]} \\
 & 2 & 88.52 & 88.37 & $-$0.15 & {\scriptsize\color{gray}[$-$0.42,\,$+$0.14]} \\
 & mean & 88.46 & 88.25 & \textbf{$-$0.21} & {\scriptsize\color{gray}[$-$0.42,\,$-$0.00]} \\
\bottomrule
\end{tabular}
\caption{\textbf{The first three-seed fine-tune on the full benchmarks.} NAVSIM: its test split; mean: the three seeds, paired.}
\label{tab:sup-remedy-bench}
\end{table}

\subsection{Stages of the training}
\label{sup:remedy-stages}
Fig.~\ref{fig:training} of the paper gives the pass rate and the paired difference of every stage and ablation.
On Odyssey most of the closed-loop step is reaching the junction: the offline stage does not reach 12.9\,\% of its scored turns, and on the turns both stages reach the step is $+$0.024 [$-$0.076, $+$0.129].
Without the speed layer the turns do not change and the AlpaSim score falls by 0.122 (Table~\ref{tab:sup-swap}).
Without the self-correction samples the policy loses turns on AlpaSim through missed turns and stalls (15 against 6\,\% of the turns), and on Odyssey in part through turns it no longer reaches.

\subsection{Further training on junction states}
\label{sup:remedy-arms}

Table~\ref{tab:sup-remedy-arms} lists every one-seed fine-tune with a full read on Odyssey and HUGSIM (the four-round fine-tune, run from three seeds, is in Table~\ref{tab:sup-remedy-seeds}); each continues seed 1 for 12{,}000 steps.
Three controls continue the same seed with no new data: for the 12{,}000 steps of the arms (`the control'), for 6{,}000 steps (`6k'), and for 6{,}000 steps with the distillation term of post-training also applied to the target-free samples, which no arm does (`earlier').
The length of the training matters on Odyssey: the control of 12{,}000 steps passes 0.092 [0.020, 0.164] more of the Odyssey turns than that of 6{,}000, while the HUGSIM and AlpaSim turns do not change resolvably.
Against the control of 12{,}000 steps no route-perturbed arm is resolved on either benchmark; against the shorter controls three of them are on Odyssey (five against the earlier control), and against the parent seed, which no choice of control touches, one.

\paragraph{One-seed fine-tunes.}
The arms of Table~\ref{tab:sup-remedy-arms} are the end of a line of 37 one-seed fine-tunes, most of them read first on small turn sets (12 Odyssey turns that one seed drives cleanly, 14 on which it runs wide, 27 AlpaSim turns with an injected push); 24 of them re-scored on five continuous measures per turn with a correction for multiple comparisons showed effects only on the injected AlpaSim pushes, where recovery data rescue 22--23 of 27 against 14 for the control.
None of the 15 pairs of the six route-perturbed arms of Sec.~\ref{sec:remedy} differs resolvably on Odyssey, and 5 do on HUGSIM, each with the half-batch arm or the six-round arm below another arm.
On the 57 AlpaSim turns no fine-tune differs resolvably from the control, and each ends more turns off the road (6--11 against 1); the recovery-data fine-tune also lowers the AlpaSim score by 0.020 [0.0002, 0.041], so its gain on the injected pushes does not reach the turns the policy drives by itself.
Its HUGSIM loss holds against all three controls; it is one arm of one seed, and the six route-perturbed arms span $-$0.166 to $-$0.003 against the control on HUGSIM, so we state the route perturbation for the family of arms and claim no single-arm effect.

\paragraph{What the recovery-data fine-tune follows.}
\label{sup:remedy-route}
Three reads stand behind the statement of Sec.~\ref{sec:remedy} that the correction learned from recovery data leans on the route; the counts on 12 and 27 turns are of turns without any incident, stricter than the pass rule, from one training run each.
Open loop, a route displaced by 1.75\,m moves the plan of the recovery-data fine-tune by 0.38 [0.32, 0.44] of the shift at 2\,s, against 0.17 [0.15, 0.19] for the control and 0.18 [0.15, 0.22] for the six-round route-perturbed fine-tune.
The route that Odyssey serves is a road-level line that lies outside the lane at 7 of the 12 corners of the 12 Odyssey turns used here; the earlier control keeps 10 of them without incident under it, the route-perturbed fine-tune 6 and the recovery-data fine-tune 2, which keeps 11 when a lane-accurate route replaces it.
Of the 27 AlpaSim turns with an injected push the two fine-tunes pass 23 and 21, the control 14.

\paragraph{The two fine-tunes run from three seeds.}
\label{sup:remedy-final}
Two route-perturbed fine-tunes were trained from all three seeds and are read against the seeds they start from (Tables~\ref{tab:sup-remedy-seeds} and~\ref{tab:sup-remedy-bench}).
The first, twelve rounds and six with kicks on the approach, was chosen after the one-seed reads, by which any of the six arms would serve on Odyssey; the second, four rounds, is the one arm resolved above the earlier control on both benchmarks among 17 one-seed fine-tunes with a HUGSIM read at two starts, and was trained again from the other two seeds for that reason.
Both gain on Odyssey over the parent seeds and against the controls of 6{,}000 steps, and neither on HUGSIM or on the AlpaSim turns; against the control of 12{,}000 steps, which exists for seed 1, neither run from that seed is resolved.
On the full benchmarks the first fine-tune changes no closed-loop score resolvably and lowers NAVSIM by 0.21 [0.00, 0.42] PDMS, most of it in seed 1, and by 0.35 [0.13, 0.57] EPDMS, through ego progress and comfort, with drivable-area compliance unchanged.
No fine-tune therefore improves turns on both benchmarks once it is read over three seeds and eight starts, and we do not attribute the Odyssey gain over the parent seeds to the route perturbation.

\paragraph{Pass rule.}
Under the other pass rules of Sec.~\ref{sup:bench-rules} (\protect\nameref{sup:bench-rules}) the Odyssey gain of the first three-seed fine-tune over its parents is $+$0.062 to $+$0.079; its interval stays above zero under four of them and touches zero when a kerb touch passes or replayed-vehicle contacts count.
\FloatBarrier
\end{document}